\documentclass[acmsmall]{acmart}

\AtBeginDocument{%
  }

\setcopyright{acmcopyright}
\copyrightyear{2025}
\acmDOI{XXXXXXX.XXXXXXX}

\setcopyright{acmlicensed}
\acmJournal{TOIS}
\acmYear{2025} \acmVolume{1} \acmNumber{1} \acmArticle{1} \acmMonth{1}\acmDOI{10.1145/3689628}

\acmJournal{JACM}
\acmVolume{37}
\acmNumber{4}
\acmMonth{8}
\usepackage{subfigure}
\usepackage{textcomp}
\usepackage{stfloats}
\usepackage{url}
\usepackage{verbatim}
\usepackage{graphicx}
\usepackage{booktabs}
\usepackage{mathrsfs}
\usepackage{mathtools}
\usepackage{color}
\usepackage{multirow}
\usepackage{adjustbox}
\usepackage{xcolor}
\usepackage{colortbl}
\usepackage{mdframed}
\usepackage{soul}

\begin{document}

\title{How Vital is the Jurisprudential Relevance: Law Article Intervened Legal Case Retrieval and Matching}

\author{Nuo Xu}
\email{nxu@sei.xjtu.edu.cn}
\orcid{1234-5678-9012}
\author{Pinghui Wang}
\authornote{Corresponding Author}
\email{phwang@mail.xjtu.edu.cn}
\affiliation{%
  \institution{MOE KLINNS Lab, Xi’an Jiaotong University}
  \city{Xi'an, Shaanxi}
  \country{China}
}


\author{Zi Liang}
\email{liangzid@stu.xjtu.edu.cn}
\affiliation{%
  \institution{MOE KLINNS Lab, Xi’an Jiaotong University}
  \city{Xi'an, Shaanxi}
  \country{China}
  }

\author{Junzhou Zhao}
\email{junzhou.zhao@xjtu.edu.cn}
\affiliation{%
  \institution{MOE KLINNS Lab, Xi’an Jiaotong University}
  \city{Xi'an, Shaanxi}
  \country{China}
}

\author{Xiaohong Guan}
\email{xhguan@xjtu.edu.cn}
\affiliation{%
  \institution{MOE KLINNS Lab, Xi’an Jiaotong University}
  \city{Xi'an, Shaanxi}
  \country{China;}
  \institution{China and Department of Automation and NLIST Lab, Tsinghua University}
  \city{Beijing}
  \country{China}
  }

\renewcommand{\shortauthors}{Xu et al.}
\newcommand{\lzm}[1]{\textcolor{red}{LZ: #1}}
\newcommand{\markred}[1]{\textcolor{red}{#1}}
\newcommand{\markblue}[1]{\textcolor{blue}{#1}}
\begin{abstract}
  Legal case retrieval (LCR) aims to automatically scour for comparable legal cases based on a given query, which is crucial for offering relevant precedents to support the judgment in intelligent legal systems.
  Due to similar goals, it is often associated with a similar case matching (LCM) task.
  To address them, a daunting challenge is assessing the uniquely defined legal-rational similarity within the judicial domain, which distinctly deviates from the semantic similarities in general text retrieval.
  Past works either tagged domain-specific factors or incorporated reference laws to capture legal-rational information.
  However, their heavy reliance on expert or unrealistic assumptions restricts their practical applicability in real-world scenarios.
  In this paper, we propose an end-to-end model named \textit{LCM-LAI} to solve the above challenges.
  Through meticulous theoretical analysis, LCM-LAI employs a dependent multi-task learning framework to capture legal-rational information within legal cases by a law article prediction (LAP) sub-task, without any additional assumptions in inference.
  Besides, LCM-LAI proposes an article-aware attention mechanism to evaluate the legal-rational similarity between across-case sentences based on law distribution, which is more effective than conventional semantic similarity.
  We perform a series of exhaustive experiments including two different tasks involving four real-world datasets. 
  Results demonstrate that LCM-LAI achieves state-of-the-art performance.
\end{abstract}




\maketitle

\newpage

\section{Introduction} \label{sec: introduction}
\begin{figure}[t]
\centering
\includegraphics[width=1.0\linewidth]{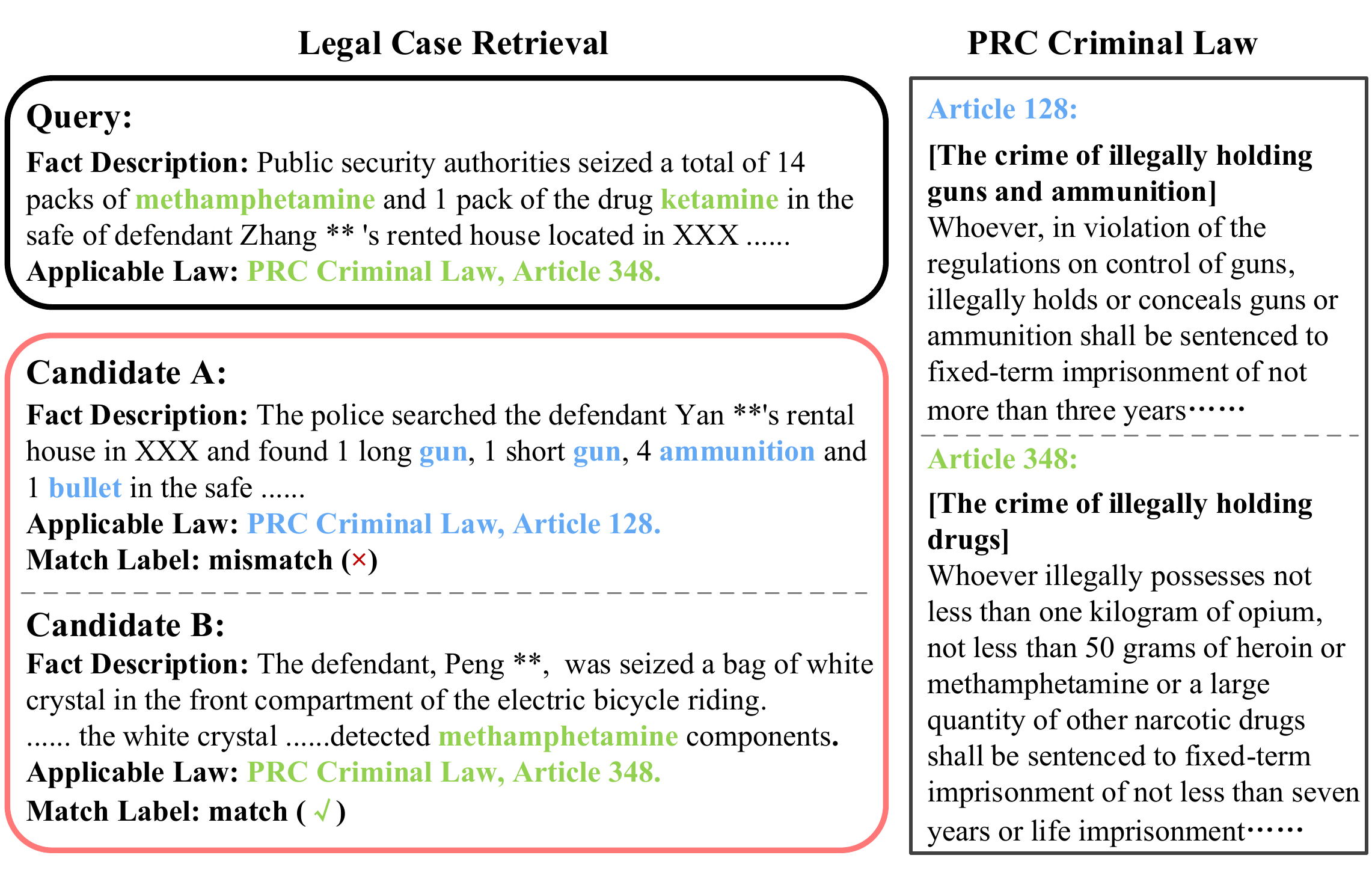}
\caption{Example of the special text relevance in LCR. Left: the query and candidate legal cases in LCR. Right: two applicable law articles of the legal cases.}
\label{fig: LCR_example}
\end{figure}
Legal case retrieval (LCR) aims to search for similar precedents for a query case, which is highly valuable in diverse judicial systems.
Because the principle of comparable judgments for similar cases
\footnote{This principle means that if different cases have the same illegal conduct, they should be sentenced to the same applicable laws, crimes, and forms of punishment (e.g., fines, detention, imprisonment, death penalty, etc). Furthermore, if the degree of illegality involved in the cases is similar, then the degree of punishment (e.g., the amount of fines, the duration of detention and imprisonment, etc.) they were sentenced should be roughly the same.} 
upholds the broad applicability and fairness of the judicial system, comparable precedents not only influence the incentives of decision-makers in the common law system but also provide the basis for legal reasoning in the civil law system.
In particular, in recent years, some countries with a civil law system (e.g., China) have promulgated official documents\footnote{https://www.court.gov.cn/fabu/xiangqing/334151.html} that institutionally incorporate the process of LCR.
This significant development has directed scholarly attention towards the challenges of LCR within the civil law system~\cite{ma2021LeCaRD, yu2022Explainable, sun2023law}.
This paper primarily investigates the LCR problem under the civil law system.


In earlier studies, researchers viewed LCR as an application of similar text retrieval models that seeks to find a precedent with a similar semantic context given a query case~\cite{xiao2021Lawformer,bhattacharya2020LCR_survey,saravanan2009LCR_1,zeng2005LCR_2,zhaochun2024answer}.
However, researchers discovered challenges that set LCR apart from general text retrievals in further research~\cite {shao2020Bert_PLI,hu2022Bert_LF}.
These challenges can be summarized as follows: (1) Excessively long legal case documents; (2) Special definition of text relevance in LCR, extending beyond the semantic similarity of conventional text retrieval; (3) Insufficient professional-labeling data for model training.
Due to the continuous release of open-source data~\cite{ma2021LeCaRD,yu2022Explainable,xiao2018CAIL2018} during the digitization of legal information and the extensive research on long-text models~\cite{Yang@Transformer_XL, NEURIPS@XL_Net, arxiv@Longformer} by NLP experts, challenges (1) and (3) have been partially overcome. 
However, challenge (2) persists, perplexing researchers, and impeding the effectiveness of current semantic-based models.
To illustrate the impact of challenge (2) on semantic-based models, we present an example related to Chinese criminal cases in Fig.~\ref{fig: LCR_example}, where we showcase two query candidates and indicate the relevant law articles\footnote{Law articles, which are usually enacted by the administration of justice (e.g., Criminal Law in China,  Act of Congress in America, Penal Code in India, and so on), are the foundation of the civil law systems. They are also termed ‘statutes’ in some countries.} on the right side. 
\begin{example}{
\emph{Candidate A} and \emph{Query} have a lot of similar text in the fact description (black font), but they apply different laws (PRC Criminal Law\footnote{https://flk.npc.gov.cn/detail2.html?ZmY4MDgxODE3OTZhNjM2YTAxNzk4MjJhMTk2NDBjOTI}, \emph{Article 128} and \emph{Article 348}) because they have different legal characteristics (green and blue font). This results in the expert-annotated labels of Query and Candidate A being "mismatch" although the existing language models calculate that they have a relatively high semantic similarity.
As for \emph{Candidate B}, it has the same applicable law articles (\emph{Article 348}) and similar legal characteristics (green font) as Query, so its label is still a "match" even though it has a lot of different contexts than Query.
}
\end{example}

To solve the challenge (2), the Chinese law system’s official guideline document\footnote{\url{https://www.court.gov.cn/fabu/xiangqing/243981.html}} highlights that a relevant case must be assessed based on three aspects, including the applicable law, the central point of the dispute, and the essential facts.
This suggests that the pertinent cases in LCR exhibit not only semantic similarities but also some level of legal congruity.
Thus, several researchers have implemented some techniques of capturing jurisprudential relevance between legal cases, which involve incorporating manual labels or elements such as legal events labels~\cite{yao2022LEVEN}, legal-rational labels~\cite{yu2022Explainable}, legal entities~\cite{shao2020Bert_PLI}, and other related factors.
Nevertheless, manually labeling these attributes can be an arduous and time-consuming process that heavily depends on experts. 
Furthermore, these labeling-based approaches tend to be highly specific to a particular legal system, consequently restricting the broad applicability of existing methodologies across diverse judicial systems.
Thus, in practical application, we desire a general and effective method to capture the jurisprudential relevance between cases.
The most relevant existing work that meets this requirement is the approach taken by Sun et al~\cite{sun2023law}, who draw inspiration from guideline documents to utilize applicable law articles to reconstruct case representations and thereby improve the overall performance of the legal case matching (LCM) task.
Their method exhibits strong generalization capabilities across criminal and civil cases, with easily accessible law articles.
Yet, despite its efficacy, the practical implementation of this approach can be challenging due to its unrealistic assumption that all query cases have access to applicable law article references. 
Actually, in the judicial practice of the civil law system, judges often use similar historical precedents to support their judgment conclusions in cases.
This means that, from the perspective of real-world judicial procedures, similar case retrieval occurs before determining the applicable law articles for cases.
Thus, using applicable law articles as input to evaluate similar cases is difficult to apply to practical scenarios.

To address the challenge of capturing jurisprudential relevance in LCR and LCM tasks under a more realistic situation assumption, in this paper, we propose an end-to-end framework named the \textbf{L}egal \textbf{C}ase \textbf{M}atching Network with \textbf{L}aw \textbf{A}rticle \textbf{I}ntervention (\textbf{LCM-LAI}). 
Regarding each law article as a text expression of jurisprudential information by professionals, LCM-LAI leverages the complete set of law articles as valuable side information.
More specifically, LCM-LAI tackles the challenges (2) mentioned above from two key aspects.
On the one hand, we treat the legal-rational information in legal case fact descriptions as intermediate variables and propose the multi-task learning framework of LCM-LAI after theoretical analysis and formula derivation (cf. Sec.~\ref{sec:analysis}).
This framework includes a relevant law article prediction sub-task to effectively capture the legal-rational information from the fact descriptions of legal cases.
Such a multi-task learning framework eliminates the strong constraint that requires applicable law articles as input like Sun et al.~\cite{sun2023law} and reasonably solves the LCR and LCM tasks with the captured legal-rational information.
On the other hand, to effectively model the jurisprudential interaction between two legal cases, LCM-LAI proposes an innovative article-aware attention mechanism that not only captures critical legal-rational information, but also measures the legal-rational correlation between two across-case sentences to enhance the model’s predictive capabilities.
Instead of directly calculating the similarity between sentence embeddings like general semantic correlation measurement, this attention mechanism composes a vector of the attention scores between sentences and each law, forming the law distribution vector. 
Subsequently, we compute the similarity between the law distribution vector of sentences as their jurisprudential interaction score.
This approach allows for a more effective measurement of the jurisprudential correlation between two cross-case sentences as it considers the distribution of legal-rational information within relevant law articles. 
Combining semantic and jurisprudential interaction knowledge, LCM-LAI outperforms existing baselines in both LCR and LCM tasks.

We summarize the contributions of our work as follows:
\begin{enumerate}
    
    \item {We introduce LCM-LAI\footnote{Our source codes are available at \url{https://github.com/prometheusXN/LCM-LAI}.}, a novel end-to-end dependent multi-task learning framework, to solve LCR and LCM tasks. Relying on a sub-task of law article prediction, it effectively evaluates the correlation between cases from both semantic and legal aspects.}
    
    \item {We propose a novel article-aware attention mechanism to effectively measure the jurisprudential correlation between two across-case sentences by considering the distribution of relevant laws instead of directly calculating the similarity between sentence embeddings.
    }
    
    \item {We conducted extensive experiments on four real-world datasets for both LCR and LCM tasks. 
    Comprehensive experiments illustrate that our model not only outperforms all existing state-of-the-art methods with a maximum of $7.02\%$ relative improvement, but also meets the time efficiency of practical applications.}
    
\end{enumerate}

We organize the rest of this paper as follows. 
Sec.~\ref{sec:related_work} summarizes related works.
Sec.~\ref{sec:methodology} formulates the problem, carries on the rigorous theoretical analysis, and obtains the framework design of the LCM-LAI.
Sec.~\ref{sec:methods} presents our method LCM-LAI in detail.
The performance evaluation and testing results are in
Sec.~\ref{sec:experiments}.
Some discussion and conclusion remarks then follow.  
\section{Related Work} \label{sec:related_work}
Our work solves the LCR \& LCM tasks through a dependent multi-task framework.
Therefore, this section introduces related research from these two aspects.



\subsection{Legal Case Retrieval and Matching}
Over the past few decades, numerous retrieval methods have been proposed, particularly for ad-hoc text retrieval.
As technology continues to develop, related methods have transitioned from the traditional bag-of-words (BOW) ones, which rely on item-matching techniques such as VSM~\cite{salton1988VSM}, BM25~\cite{robertson1994BM25}, and LMIR~\cite{song1999LMIR}, to machine learning methods that utilize artificial features, such as LTR~\cite{liu2009LTR}.
Most recently, data-driven deep learning methods, such as DSSM~\cite{hu2014DSSM}, MatchPyramid~\cite{pang2016MatchPyramid}, Match-SRNN~\cite{wan2016Match-SRNN} and others, have emerged as a leading approach.
These studies have also demonstrated remarkable success in practical applications, including website search engines, recommendation systems, etc.

However, legal case retrieval and matching, being the critical applications of the retrieval model in the legal field, pose several unique challenges in comparison to general ad-hoc text retrieval. These challenges include specialized definitions of relevance in law~\cite{van2017SpecialDefinition}, dealing with extremely lengthy documents, and handling highly professional expression~\cite{turtle1995Legalworld}.
Consequently, in earlier times, certain solutions that heavily relied on expert knowledge were proposed to evaluate the legal relevance between two legal documents, such as ontological frameworks~\cite{saravanan2009LCR_1} and legal issue decomposition~\cite{zeng2005LCR_2}.
With the advancement of NLP technology, particularly since the introduction of pre-training language models (PLMs) like BERT~\cite{devlin2018bert} was proposed, researchers try to employ such data-driven deep learning methods to solve LCM \& LCR tasks.
Unfortunately, because the length of legal documents far exceeds the input length limit of PLMs like BERT, the straightforward truncation strategy adopted by popular passage retrieval methods (e.g., DPR~\cite{karpukhin2020DPR}, ColBERT~\cite{khattab2020colbert}, etc.) in solving LCM \& LCR tasks will lose a lot of the case information, resulting in poor performance.
Therefore, when dealing with the LCR \& LCM tasks, existing approaches typically segment legal documents into paragraphs or sentences and capture the interaction at the paragraph or sentence level between cases to assess their relevance. 
In this regard, one of the most typical works is proposed by Shao et al.~\cite{shao2020Bert_PLI}, which employs BERT to capture the paragraph-level semantic interaction information, which overcomes the challenge of handling lengthy documents in the LCR \& LCM tasks.
However, their work only assesses the relevance of legal cases at a semantic level and makes some oversights due to neglecting the legal correlation between them.
To address this issue and capture the legal relevance between cases, some works have begun to introduce additional expertise to assist with the task of LCR \& LCM.
Hu et al.~\cite{hu2022Bert_LF} strengthen the retrieval performance of Shao et al.'s model in criminal cases by appropriately labeling and encoding particular legal entities associated with the PRC’s Criminal Law.
Yao et al.~\cite{yao2022LEVEN} effectively enhance the legally-relevant statements of the fact description by defining and labeling a vast array of legal events, predicting their corresponding labels, and subsequently supplementing each event with additional embeddings.
Yu et al.~\cite{yu2022Explainable} advance the performance of the LCM task by not only labeling the legal rationale statement of each case but also identifying the alignment relationship between the rationales in each pairwise case.
Besides, they propose an explainable model that accurately captures the legal relevance between cases by predicting the labels as mentioned above.
To the best of our knowledge, Sun et al.~\cite{sun2023law} is the first to utilize the law articles as a supportive tool for the LCM task.
Unlike previous works that heavily relied on expert knowledge annotation, they propose a causal learning approach to boost the performance of the LCM task. 
This approach involves using the applicable law as an intermediate variable, enabling the extraction of legal-related features of cases by reconstructing case representations.
However, their assumption that utilizes ground-truth applicable law articles as input limits the practicality of their work. 
As the ultimate objective of the LCM task is to achieve judicial decision-making (e.g., applicable law articles, charges, and terms of penalty), applicable law may not always be readily available.

Unlike them, our proposed LCM-LAI introduces an end-to-end dependent multi-task framework that effectively captures legal relevance between cases.
Our framework indirectly incorporates legal knowledge from law articles into the core LCR \& LCM task by the subtask of law article prediction, which not only guarantees consistency between training and practical application but also reduces reliance on experts.

\subsection{Multi-task Learning in Legal Domain}
Multi-task learning (MTL) aims to improve the performance of multiple relevant tasks simultaneously by exploiting their internal relations.
In NLP tasks, knowledge transfer between relevant subtasks is typically accomplished by sharing representations~\cite{liu2015SharingRep} or parameters~\cite{liu2018SharingParameter}, as demonstrated by much of the current research.
For example, Dong et al.~\cite{dong2015MultiTranlate} demonstrate the effectiveness of sharing encoders to enhance one-to-many neural machine translations.
Firat et al.~\cite{firat2016multi} improve multi-way and multilingual machine translation through a sharing attention mechanism.
Guo et al.~\cite{guo2018soft} propose a multi-task learning framework to incorporate auxiliary tasks of question and entailment generation, thereby enhancing the capacity for abstractive summarization.
In the legal domain, much of the current work centers around the design of dependencies between sub-tasks, with a primary emphasis on legal judgment prediction (LJP) tasks of single legal case~\cite{zhong2018Topology,yang2019Bi-feedback,yue2021Neurjudge,lyu2022CEEN,zhaochun2023multi,zhaochun2023syllogistic,LJP@TOIS2024_XN}.
Zhong et al.~\cite{zhong2018Topology} first explore the three subtasks of LJP and propose a directed acyclic graph-based topological multi-task learning framework to model their dependency.
Subsequently, Yang et al.~\cite{yang2019Bi-feedback} augment the framework as mentioned above by incorporating the backward dependency between these subtasks, further improving its efficacy.
Furthermore, Yue et al.~\cite{yue2021Neurjudge} improve the performance of multi-task learning on JLP tasks by disentangling the input fact description into fine-grained representations associated with each subtask, thereby enhancing the performance of their proposed model.

However, the above multi-task learning frameworks are all designed heuristically according to the actual judging process.
Different from them, we combine mediation analyses and naive Bayesian theory to derive and design our dependent multi-task learning framework, which achieves significant performance improvement in our experiments.
Meanwhile, to the best of our knowledge, we are the first to use the multi-task framework to improve the LCR \& LCM main task, which involves pairwise legal cases.
\section{Methodology} \label{sec:methodology}
\begin{table*}[t]
\caption{Main mathematical notations.}\label{tab:Notation}
\centering
\resizebox{\textwidth}{!}{%
\begin{tabular}{c|c}
\toprule 
Notation & Description\\
\midrule 

$\mathcal{L}\triangleq\{L_1, \ldots, L_{n_L}\}$ &
the set of all involved law articles \\

$X = \{x_1, \ldots, x_{n_x}\}$ and $Y = \{y_1, \ldots, y_{n_y}\} $ & 
the sentences sequence of query case $X$ and candidate case $Y$ \\

$X_L \subseteq X$ and $Y_L \subseteq Y$ & 
the set of legal-rational relevant sentences in cases $X$ and $Y$ \\

$\mathcal{L}_X, \mathcal{L}_Y \subseteq \mathcal{L} $ & 
the case $X$ and $Y$'s corresponding applicable law article set\\

$R = \langle X, Y \rangle$ & 
the simplified notation for a case pair containing $X$ and $Y$ \\

$R_{L} = \langle X_L, X_L \rangle $ &
the simplified notation for the set pair containing $X_L$ and $Y_L$ \\

$\mathcal{L}_{R} = \langle \mathcal{L}_X, \mathcal{L}_Y \rangle$ &
the simplified notation for a pair of applicable law article sets containing $\mathcal{L}_X$ and $\mathcal{L}_Y$ \\

$\mathbf{x}_i$, $\mathbf{y}_j$, and $\mathbf{L}_k$ & 
the encoded sentence embedding of sentence $x_i \in X $, $y_j \in Y$ and law article $L_k \in \mathcal{L}$ \\

$\textbf{Z}_{L}^{X} = [z_{1}^{X}, \ldots, z_{n_L}^{X}]$ and $ \textbf{Z}_{L}^{Y} = [z_{1}^{Y}, \ldots, z_{n_L}^{Y}]$ &
the law article label vector of case $X$ and $Y$, where $z_{k}^{*} \in \{0, 1\}$ \\

$\textbf{Z}(X, Y) = [z_1, \ldots, z_{|Z|}]$ & 
the one-hot ground-truth LCM vector of legal case pair $(X, Y)$\\

$S(X,Y) \in [0, 1]$ & 
the matching score between query $X$ and candidate $Y$\\

$\mathcal{M} = \{ \textbf{m}_1, \ldots, \textbf{m}_{n_L}\}$ & 
the set for memory vectors of law articles\\

$\mathbf{C}^{(S)}, \mathbf{C}^{(L)} \in \mathbb{R}^{n_x \times n_y}$ & 
the semantic and legal correlation matrix for each pairwise cross-case sentence \\

$\mathbf{X}^{(S)}, \mathbf{Y}^{(S)}$ &
the semantic interaction representations of case $X$ and case $Y$\\

$\mathbf{X}^{(L)}, \mathbf{Y}^{(L)}$ &
the legal interaction representations of case $X$ and case $Y$\\

$\mathbf{X}^{(A)}, \mathbf{Y}^{(A)}$ &
the article-intervened representations of case $X$ and case $Y$\\

$\mathbf{X}_f, \mathbf{Y}_f$ &
the final interaction representations of case $X$ and case $Y$\\

\bottomrule
\end{tabular}%
}
\end{table*}
In this section, we first formalize the LCR and LCM tasks. 
Subsequently, we employ the mediation variable analysis method and naive Bayes theory to analyze the problem, culminating in the framework of our LCM-LAI based on our analysis conclusion.

\subsection{Problem Formulation}
Here, we first show some notations in Tab.~\ref{tab:Notation} and then formulate the LCR and the LCM tasks in the following.

\noindent {\bf Legal Cases and Law Articles.} 
As our method tries to add the law articles as the external knowledge to improve the performance of LCR tasks, we first formalize the legal cases and law articles here.
In our work, we treat each legal case $X$ as a sequence of sentences, i.e., $X = \{ x_1, \ldots, x_{n_x}\}$, where $x_i$ denoted the $i$-th sentence of case $X$, and $n_x$ represents the number of sentences in the case $X$.
In civil law countries, legal cases are often analyzed and adjudicated according to statutory law (also known as written law).
We denote the statutory law as a set of law articles $\mathcal{L}\triangleq\{L_1, \ldots, L_{n_L}\}$, where $n_L$ is the article number of the full law article set. 
Since law articles' length is generally short, we represent each law article $L_i$ as a sentence.
Besides, for each legal case $X$, we also denote its cited law article set with $\mathcal{L}_{X}$, where $\mathcal{L}_{X} \subseteq \mathcal{L}$ and $\|\mathcal{L}_{X}\| > 0$.

\noindent {\bf Legal Case Retrieval.}
Formally, given a query case $X$, and a set of candidate cases $\mathcal{Y} = \{Y_1, \ldots, Y_n\}$, 
the goal of the LCR task is to identify the relevant and instructive cases $\mathcal{Y}^{*} = \{Y_i^{*} | Y_i^{*}\in \mathcal{Y} \land \mathit{related}(Y_i^{*}, X) \}$, where $\mathit{related}(Y_i^{*}, X)$ denotes that $Y_i^{*}$ is related to the given query case $X$.
To evaluate the relevance between the query and candidates, existing methods generally learn a matching model $ \mathrm{F}(X, Y_i) \rightarrow S_i $, where the higher output score $S_i$ indicates the stronger correlation between the query case $X$ and candidate case $Y_i$.

\noindent {\bf Legal Case Matching.}
As a critical step for the LCR task, we also formalize the LCM task here.
Given a set of labeled data tuples $\mathcal{D}\triangleq\{(X, Y, z)_{i}\}_{i=1}^{N}$, where $X$ and $Y$ denote two source legal cases, $z \in Z$ denotes the human-annotated matching label and $N$ denotes the tuple number in the set.
$Z$ could be the manually defined match levels.
For example, when $Z = \{0, 1, 2\}$, the values $0$, $1$, and $2$ correspond to a mismatch, a partial match, and a complete match, respectively.
The LCM task aims to learn a matching model $\mathrm{F}(X, Y) \rightarrow Z$ to predict the matching label $\hat{z} \in Z$ by inputting two legal cases $X$ and $Y$, which is trained based on the labeled tuples in dataset $\mathcal{D}$.

Since the forms are similar and for the convenience of subsequent discussion, we unify the goal of LCR \& LCM tasks, i.e., $\mathrm{F}(X, Y) \rightarrow Z$, where the only difference is that the label field $Z$ of LCR task is continuous while that of LCM task is discrete.

\begin{figure}[t]
\centering
\includegraphics[width=0.70\linewidth]{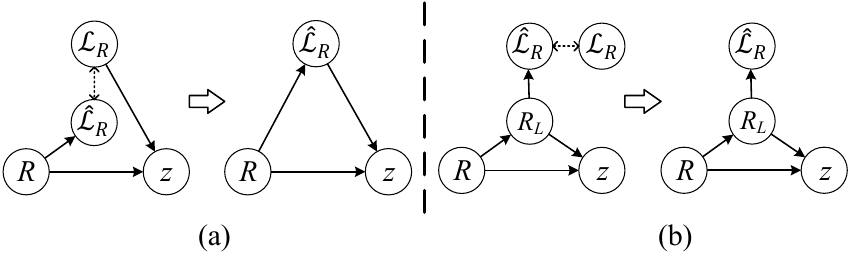}
\caption{The causal graph of two DMT frameworks. (a) The inconsistency between the training and reasoning stages leads to the train-test discrepancy. (b) The potential intermediate variable $R_L$ is introduced, making the training and reasoning process consistent.}
\label{fig: problem_analyse}
\end{figure}

\subsection{Problem Analysis}\label{sec:analysis}
According to the above definition, given a labeled sample $(X, Y, z)$, the objective of a desired case matching model is to maximize the prediction probability $\mathrm{P}(z|X, Y)$.
As mentioned in Sec.~\ref{sec: introduction}, relevant cases should be assessed from both semantic and jurisprudential relevance.
Thus, following the official guide document (referenced in Sec.~\ref{sec: introduction}), we introduce the applicable law articles of cases to assess the jurisprudential relevance between them.
Take $\mathcal{L}_X$ and $\mathcal{L}_Y$ as the applicable law articles of case $X$ and $Y$ respectively, 
a straightforward way to model LCR \& LCM tasks is to maximize the objective $\mathrm{P}(z|\mathcal{L}_X, \mathcal{L}_Y, X, Y)$, similar to the method employed by Sun et al.~\cite{sun2023law}.
However, such a model ignores the reality that in the actual judicial case hearing process, the judgment of the applicable law occurs after the retrieval of similar cases, making it difficult to apply such models in practice.
Fortunately, inspired by related work on legal judgment prediction~\cite{xu2020LADAN,LJP@TOIS2024_XN,zhong2018Topology,yue2021Neurjudge}, we find that the definition of applicable law articles serves as a high-dimensional summary of the legal-ration information contained in the fact description of the corresponding legal cases.
Thus, to capture the jurisprudential relevance between legal cases, our method treats the applicable law articles as the intermediate variables and forms the LCR \& LCM task as a dependent multi-task (DMT) learning problem.
From this perspective, the most direct modeling approach is to utilize the case facts to predict the corresponding applicable law articles (whose objectives are $\mathrm{P}(\mathcal{L}_X|X)$ and $\mathrm{P}(\mathcal{L}_Y|Y)$), and then use the case facts and the predicted applicable law articles to jointly predict the matching relationship between cases (whose objective is $\mathrm{P}(z|\mathcal{L}_X, \mathcal{L}_Y, X, Y)$).
Thus, such a model needs to maximize the following overall objective,
$$
\mathrm{P}(z, \mathcal{L}_X, \mathcal{L}_Y | X, Y ) = \mathrm{P}(\mathcal{L}_X|X) \cdot \mathrm{P}(\mathcal{L}_Y|Y) 
\cdot \mathrm{P}(z| \mathcal{L}_X, \mathcal{L}_Y,  X, Y),
$$
To simplify the following formulas, we simplify the case pair $\langle X, Y \rangle$\footnote{Note that in this paper, $<\cdot, \cdot>$ only represents the tuple relationship instead of any computational operation.} as $R$ 
and simplify the pair of the corresponding applicable law article set $\langle \mathcal{L}_X, \mathcal{L}_Y \rangle$ as $\mathcal{L}_R$.
Thus, we get a simplified DMT formula as follows
\footnote{Considering $\mathcal{L}_X$ and $\mathcal{L}_Y$ are two independent variables, and case $X$ and $Y$ are also independent, we get $\mathrm{P}(\mathcal{L}_X|X) \cdot \mathrm{P}(\mathcal{L}_Y|Y)=\mathrm{P}(\mathcal{L}_X, \mathcal{L}_Y|X, Y)=\mathrm{P}(\mathcal{L}_R|R)$.},
\begin{equation} \label{eq: original model}
    \mathrm{P}(z, \mathcal{L}_R|R) = \mathrm{P}(\mathcal{L}_R|R) \cdot \mathrm{P}(z|\mathcal{L}_R, R).
\end{equation}
However, there is a train-test discrepancy in Eq.~\ref{eq: original model} in the causal view. 
The causal graph of such a framework is shown in Fig~\ref{fig: problem_analyse}(a).
During the training stage (the left half of Fig.~\ref{fig: problem_analyse}(a)), we can use the gold label of applicable law articles $\mathcal{L}_R$ to pre-train the object $P(z|\mathcal{L}_R, R)$, while in the inference stage (the right half of Fig.~\ref{fig: problem_analyse}(a)), we can only use the predicted label $\hat{\mathcal{L}_R}$ to predict the matching result, as the query cases are typically pending adjudication (i.e., lacking explicitly applicable law articles).
Referring to previous works~\cite{chen2020Causal_1, chen2021Causal_2}, such a discrepancy would hurt the inference performance.

To overcome such a train-test discrepancy, we reanalyze the DMT learning in LCR \& LCM tasks. 
As mentioned in Sec. \ref{sec: introduction} and Fig.~\ref{fig: LCR_example}, for each legal case, there is always some legal-rational part of the fact description that corresponds to its applicable law articles. 
Designating this legal-rational portion as $R_L$, our focus lies in isolating this element within case fact descriptions and determining its relevance across cases.
Thus, we can update Eq.~\ref{eq: original model} as,
\begin{equation} \label{eq: new model}
P(z, R_L|R) = P(R_L|R) \cdot P(z|R_L, R).  
\end{equation}
However, it is not easy for a deep-learning model to separate the legal-rational part $R_L$ directly from the input $R$ due to the non-differentiability of the operation.
Inspired by previous works~\cite{xu2020LADAN, luo2017FLA}, the law article prediction task based on the attention mechanism can well extract the legal-rational part's vector representation in the vector space. 
Therefore, we tend to inject a term of legal prediction task $P(\mathcal{L}_R|R)$ into Eq.~\ref{eq: new model}.
In addition, it's worth noting that $R_L$ ideally has two particular properties, i.e., $R_L \in R$ and $\mathcal{L}_R$ is only relevant to $R_L$. 
Thus, we approximate two formulas by these two properties, i.e., $ P(\mathcal{L}_R| R_L, R) \geq P(\mathcal{L}_R|R) $ and $ P(\mathcal{L}_R|R_L) \approx \alpha$. 
In this way, we further deduce the formula,
\begin{equation} \label{eq: final model}
\begin{aligned}
P(z, R_L|R) &= P(R_L|R) \cdot P(z|R_L, R)\\
&=\frac{P(\mathcal{L}_R|R_L, R)}{P(\mathcal{L}_R|R_L)} \cdot P(z|R_L, R)\\
&\geq \frac{1}{\alpha} P(\mathcal{L}_R|R) \cdot P(z|R_L, R).
\end{aligned}
\end{equation}

So far, the two terms of Eq.~\ref{eq: final model} inspire us that the reasonable model design has two necessary components: the first term $P(\mathcal{L}_R|R)$ indicates the subtask of law article prediction, and the second term $P(z|R_L, R)$ states that the model needs to combine the original input $R$ and the vector representation $R_L$ of the legal-rational part to predict the results of LCR \& LCM tasks.

\begin{figure*}[t]
\centering
\includegraphics[width=1.0\linewidth]{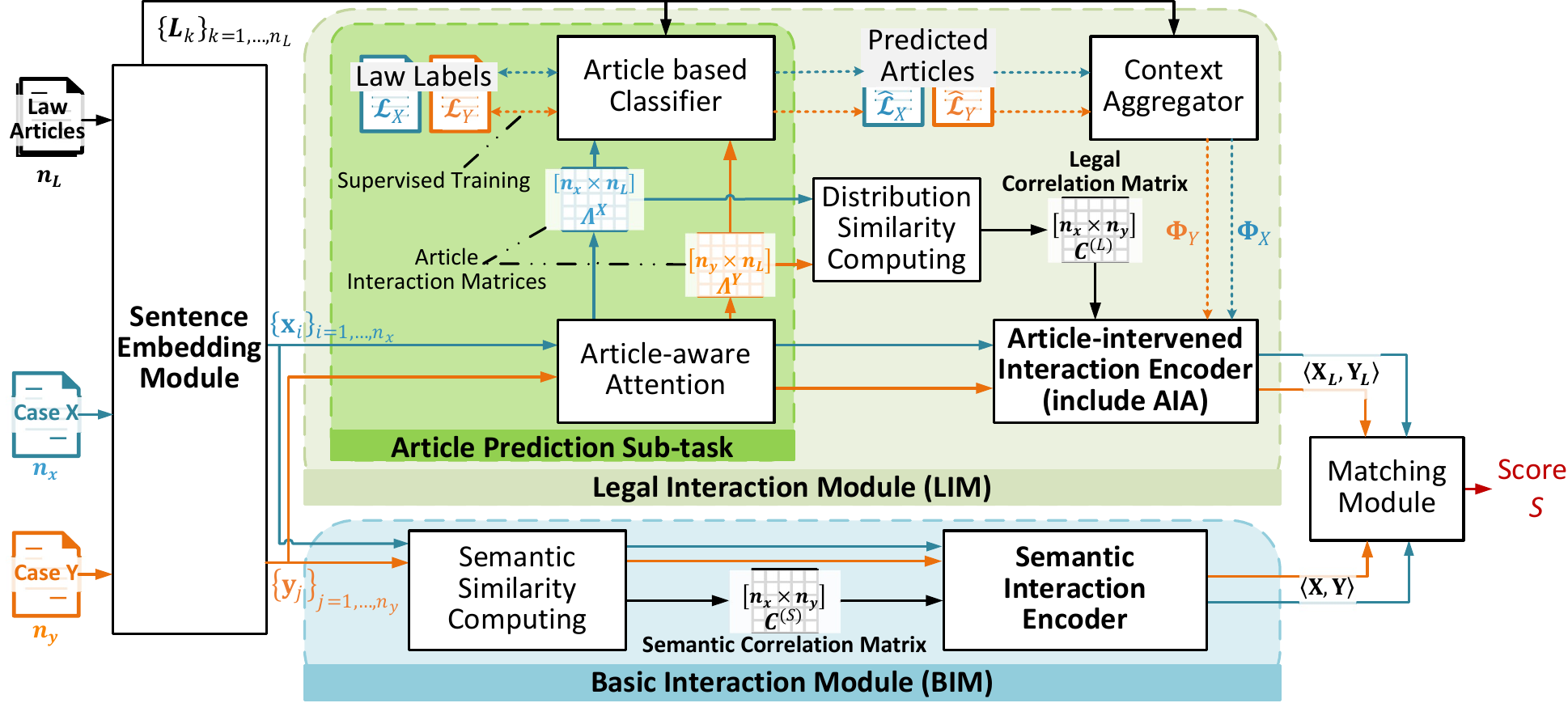}
\caption{
Overview of our framework LCM-LAI. 
This framework operates by taking textual descriptions of legal cases and text definitions of law articles as inputs. 
On the one hand, it leverages the \textbf{Basic Interaction Module (BIM)} to extract the semantic interaction information between cases based on the semantic similarity between across-case sentences. 
On the other hand, it uses the \textbf{Legal Interaction Module (LIM)} to capture the legal interaction information from the perspective of the similarity of legal distribution, in which the \textbf{article prediction sub-task} is introduced to capture the legal distribution of each sentence.
Besides, LIM uses an \textbf{article-intervened attention (AIA)} mechanism to highlight the key jurisprudence-related parts of cases that rely on the predicted law articles.
Finally, LCM-LAI combines the semantic and legal interaction representations to compute the matching score.
}
\label{fig: framework}
\end{figure*}

\subsection{Framework of LCM-LAI} \label{subsec: framework}
When we substitute the simplifications, i.e., $R \rightarrow \langle X, Y \rangle$, $R_L \rightarrow \langle X_L, Y_L \rangle$, and $\mathcal{L}_R \rightarrow \langle \mathcal{L}_X, \mathcal{L}_Y \rangle$ into Eq.~\ref{eq: final model}, then we get the following formula,
\begin{equation} \label{eq: model framework}
\mathrm{P}(z, X_L, Y_L | X, Y ) = \mathrm{P}(\mathcal{L}_X|X) \cdot \mathrm{P}(\mathcal{L}_Y|Y) 
\cdot \mathrm{P}(z| X, Y, X_L, Y_L).
\end{equation}
This formula inspires the framework of our LCM-LAI model, which includes: the inspiration of the term $\mathrm{P}(z|X, Y, X_L, Y_L)$ is to extract the features from both semantic (i.e., $X, Y$) and legal-rational (i.e., $X_L, Y_L$) levels to model the interaction between two cases. The inspiration of the term $\mathrm{P}(\mathcal{L}_X|X)$ and $\mathrm{P}(\mathcal{L}_Y|Y)$ is to use the applicable law article prediction sub-task to make the model learn how to extract the legal-rational features in the case.

Thus, the framework of our LCM-LAI is shown in Fig.~\ref{fig: framework}, which can be divided into four main modules, i.e., sentence embedding module, basic interaction module (BIM), legal interaction module (LIM), and matching module.
The sentence embedding module encodes the sentences of cases' fact descriptions into vector representations.
Subsequently, when capturing the interaction information between legal cases, our LCM-LAI employs two parallel interaction modules, namely the basic interaction module and the legal interaction module, to effectively model the paired terms $\langle X, Y \rangle$ and $\langle X_L, Y_L \rangle$ in Eq.~\ref{eq: model framework}.
On the one hand, the BIM computes the basic interaction representations $\langle \mathbf{X}, \mathbf{Y} \rangle$, which contain the general semantic matching information between two legal cases, just like the conventional text matching models work.
On the other hand, the LIM generates the legal interaction representations $\langle \mathbf{X}_L, \mathbf{Y}_L \rangle$, which capture not only the legal information of each case but also the matching information from a legal perspective.
Besides, to satisfy the formula terms $\mathrm{P}(\mathcal{L}_X|X)$ and $\mathrm{P}(\mathcal{L}_Y|Y)$, we also construct the law article prediction sub-task (i.e., the Article Prediction Sub-task in Fig.~\ref{fig: framework}) in the LIM, which also helps capture the legal interaction between cases effectively.
Finally, the $\langle \mathbf{X}, \mathbf{Y} \rangle$ and $\langle \mathbf{X}_L, \mathbf{Y}_L \rangle$ are fed into the subsequent matching module to predict the matching score or label of the subsequent LCR \& LCM tasks.

Note that the core of our framework is to learn how to capture key legal-rational features under the supervision of historical precedents' applicable law article labels during training. 
As for the inference stage, our framework only focuses on the legal-rational information extracted by the model and the legal interaction between cases instead of taking the prediction results of the applicable law as the evaluation basis.
Thus, such a framework does not contravene the judicial practice procedure of the civil law system, in which the judicial judgment conclusions should be made after the similar case retrieval.

\section{Model Detail} \label{sec:methods}
In this section, we introduce the details of our LCM-LAI to implement the final learning objective mentioned above, which can be divided into three main parts: \textbf{sentence embedding module}, \textbf{basic interaction module} (\textbf{BIM}), and \textbf{legal interaction module} (\textbf{LIM}).

\subsection{Sentence Embedding Module}
Firstly, we try to encode each legal case sentence to an embedding because our LCM-LAI models the interaction between legal cases from the sentence level.
In practice, we follow the typical method mentioned in~\cite{yu2022Explainable} that uses the output of the [CLS] token of a BERT model pre-trained on a Chinese legal case corpus\footnote{The pre-trained models we used are available at \url{https://github.com/thunlp/OpenCLaP}. 
Notice, that LCM-LAI also applies to the other pre-train language models with the corresponding embeddings.} as the sentence embedding. 
In the same way, we also encode legal articles because they are generally short enough to be treated as sentences.
For the specific sentences $x_i\in X$, $y_j\in Y$, and the specific law article $L_k\in \mathcal{L}$, we formalize the calculation process as follows,
$$
\mathbf{x}_i = \text{BERT}(x_i)[\textbf{CLS}], \ \ i \in \{1, \ldots, n_x\}
$$
$$
\mathbf{y}_j = \text{BERT}(y_j)[\textbf{CLS}], \ \ j \in \{1, \ldots, n_y\}
$$
$$
\mathbf{L}_k = \text{BERT}(L_k)[\textbf{CLS}], \ \ k \in \{1, \ldots, n_L\},
$$
where $\mathbf{x}_i, \mathbf{y}_i, \mathbf{L}_k \in \mathbb{R}^{d_b}$ are the corresponding sentence embeddings.

\subsection{Basic Interaction Module}
Referring to Sec.~\ref{subsec: framework}, the two interaction information $\langle \mathbf{X}, \mathbf{Y} \rangle$ and $\langle \mathbf{X}_L, \mathbf{Y}_L \rangle$ are both critical to the LCR \& LCM tasks.
To decouple $\langle \mathbf{X}, \mathbf{Y} \rangle$ with the legal-rational interaction $\langle \mathbf{X}_L, \mathbf{Y}_L \rangle$ that we emphasize, in the BIM, we understand and model it from the perspective of general semantic similarity following the consistent idea of general text matching~\cite{yu2022Explainable, reimers2019sentence-BERT, yin2016ABCNN}.

\subsubsection{Semantic Correlation Matrix.} \label{subsubsec: semantic correlation matrix}
From a sentence-level perspective to capture the semantic matching information between legal cases, we construct a semantic matrix $\mathbf{C}^{(S)} \in \mathbb{R}^{n_x \times n_y}$ to indicate the semantic correlation between cross-case sentence pairs.
Inspired by the process in~\cite{yu2022Explainable}, we choose the euclidean-distance-based similarity to measure used to evaluate the semantic similarity between two sentences, i.e., $c_{i, j}^{(S)} = \text{dis-sim}(\mathbf{x}_i, \mathbf{y}_j)$, where $\text{dis-sim}(\cdot)$ denotes the euclidean-distance-based similarity computation function.
As with most multi-task frameworks, to improve the robustness of the model, we map the input sentence embeddings with a trainable two-layer MLP.
The specific formula is,
\begin{equation} \label{eq: semantic correlation matrix}
    c^{(S)}_{i,j} = -||\text{MLP}(\mathbf{x}_i) - \text{MLP}(\mathbf{y}_j)||_{2}=
    -\sqrt{(\text{MLP}(\mathbf{x}_i) - \text{MLP}(\mathbf{y}_j))^\mathsf{T}(\text{MLP}(\mathbf{x}_i) - \text{MLP}(\mathbf{y}_j))},
\end{equation}
where $\text{MLP}(\cdot)$ denotes the MLP operator, $||\cdot||_{2}$ denotes the euclidean norm and $c^{(S)}_{i,j} \in \mathbf{C}^{(S)}$.
\subsubsection{Semantic Interaction Encoder.} \label{subsubsec: sentence interaction}
To achieve fine-grained extraction of semantic matching information between legal cases, our first step is to model the interaction between each cross-case sentence pair to capture their matching behavior.
To be precise, in the legal case pair $(X, Y)$, we begin by using a softmax function to calculate the semantic interaction weight $\alpha_{i,j}^{(S)}$  (resp. $\beta_{i,j}^{(S)}$) for each sentence $\textbf{x}_i$ (resp. $\textbf{y}_j$) in response to the corresponding row $c_{i, :}^{(S)} $ (resp. column vector $c_{:, j}^{(S)}$) in the semantic correlation matrix $\mathbf{C}^{(S)}$ about the other legal case, i.e., 
$\alpha_{i,:}^{(S)} = \text{softmax}\left(c_{i, :}^{(S)}\right)$ and $\beta_{:,j}^{(S)} = \text{softmax}\left(c_{:, j}^{(S)}\right)$.
To show this in more detail, we provide the following element-wise expansion formula,
\begin{equation} \label{eq: Semantic Interaction Attention}
\alpha_{i,j}^{(S)} = \frac{\exp(c_{i,j}^{(S)})}
{\sum_{j=1}^{n_y} \exp(c_{i, j}^{(S)})}; \ \ 
\beta_{i,j}^{(S)} = \frac{\exp(c_{i,j}^{(S)})}
{\sum_{i=1}^{n_x} \exp(c_{i, j}^{(S)})}, \ \ 
\end{equation}
Then, we weighted aggregate all sentences in another legal case to update the new representation after semantic interaction, that is,
\[
\mathbf{x}_i^{(S)} = \mathbf{x}_i \oplus \sum_{j=1}^{n_y}\alpha_{i, j}^{(S)}\mathbf{y}_j;
\ \ 
\mathbf{y}_j^{(S)} = \mathbf{y}_j \oplus \sum_{i=1}^{n_x}\beta_{i, j}^{(S)}\mathbf{x}_i,
\]
where $\mathbf{x}_i^{(S)}, \mathbf{y}_j^{(S)} \in \mathbb{R}^{2d_b}$ denote the semantic interaction sentence representations and $\oplus$ denotes the concatenation operator.

Then, referring to the practice of Shao et al~\cite{shao2020Bert_PLI}, we also leverage a bidirectional GRU (Bi-GRU) model to further aggregate contextual information of each case, i.e., 
\begin{equation} \label{eq: Semantic RNN}
\begin{split}
& \mathbf{H}_{X}^{(S)} = [\mathbf{h}_{x, 1}^{(S)}, \cdots, \mathbf{h}_{x, n_x}^{(S)}] = \text{Bi-GRU}([\mathbf{x}_1^{(S)}, \mathbf{x}_2^{(S)}, \cdots, \mathbf{x}_{n_x}^{(S)}]) \\ 
& \mathbf{H}_{Y}^{(S)} = [\mathbf{h}_{y, 1}^{(S)}, \cdots, \mathbf{h}_{y, n_y}^{(S)}] = \text{Bi-GRU}([\mathbf{y}_1^{(S)}, \mathbf{y}_2^{(S)}, \cdots, \mathbf{y}_{n_y}^{(S)}]),
\end{split}
\end{equation}
where $\mathbf{h}_{x, i}^{(S)}, \mathbf{h}_{y, j}^{(S)} \in \mathbb{R}^{d_s}$ is the hidden state of input sentence embeddings $\mathbf{x}_i^{(S)}$ and $\mathbf{y}_j^{(S)}$, and $d_s$ denotes the dimension of legal cases' semantic interaction representations.
Finally, we use the element maximum pooling operator to obtain the semantic interaction representation of each legal case, that is,
\begin{equation} \label{eq: Semantic Representation}
\mathbf{X}^{(S)} = \text{Max-pooing}(\mathbf{H}_{X}^{(S)});
\ \ 
\mathbf{Y}^{(S)} = \text{Max-pooing}(\mathbf{H}_{Y}^{(S)}),
\end{equation}
where $\mathbf{X}^{(S)}, \mathbf{Y}^{(S)} \in \mathbb{R}^{d_s}$ separately denote the semantic interaction representation of cases $X$ and $Y$.

\subsection{Legal Interaction Module}
\begin{figure*}[t]
\centering
\includegraphics[width=1.0\linewidth]{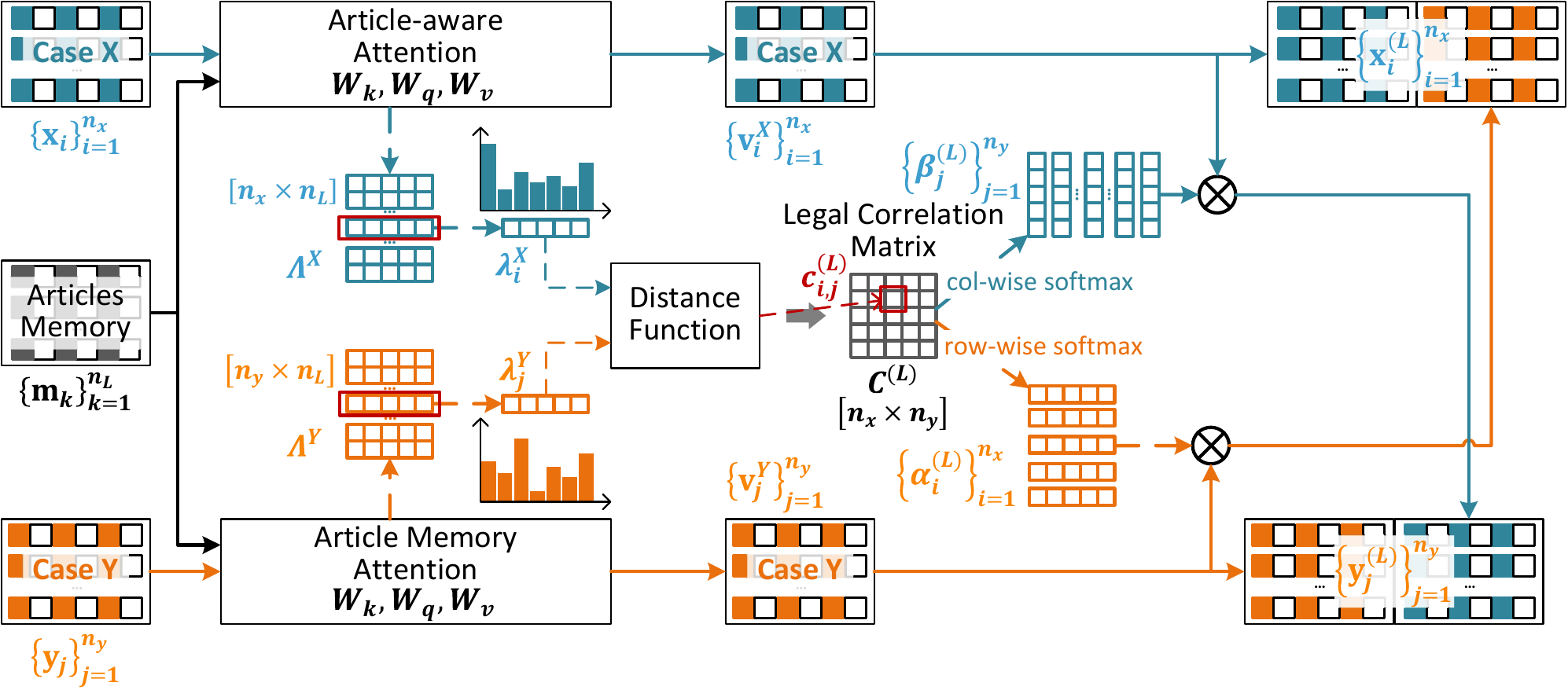}
\caption{
Body Operations of the Legal Interaction Module.
It cleverly exploits the intermediate representations of the attention mechanism of the law prediction sub-task.
It not only takes the inner value vectors to compute the legal representations but also innovatively regards the attention score vectors of law prediction as the law article distribution for computing the legal-rational correlation between sentences across cases.
}
\label{fig: Law Interaction}
\end{figure*}

To capture the legal interaction information $\langle \mathbf{X}_L, \mathbf{Y}_L \rangle$ of a legal case pair, the strategy of LIM involves two key steps. First, we capture the legal-rational information of each case through the law article prediction sub-task (cf. Sec.~\ref{subsubsec: article prediction subtask}).
Next, with the assistance of the law-distribution-aware correlation matrix (cf. Sec.~\ref{subsubsec: law correlation matrix}), we use the proposed article-intervened interaction encoder (cf. Sec.~\ref{subsubsec: law interaction encoder}) to model the legal interaction between cross-case sentence pairs.
For ease of understanding, we have shown the core of the legal interaction module in Fig.~\ref{fig: Law Interaction}.

\subsubsection{Article Prediction Sub-task.}\label{subsubsec: article prediction subtask}
Since a legal case may be associated with multiple law articles, each indicating different corresponding jurisprudential attributes, a multi-label classification approach is adopted for the article prediction sub-task, in contrast to the single-label classification method more commonly utilized in LJP. 
This enables us to capture a comprehensive range of jurisprudence-related information in a legal case.

Take the case $X$ as an example, we employ a Bi-GRU to capture contextual information for each sentence $x_i$, that is,
\[
[\mathbf{h}_1, \cdots, \mathbf{h}_{n_x}] = \text{Bi-GRU}([\mathbf{x}_1, \cdots, \mathbf{x}_{n_x}]),
\]
where $\mathbf{h}_i \in \mathbb{R}^{d_b}$ denotes the corresponding hidden state of the input sentence embedding $\mathbf{x}_i$.

\noindent {\bf Article-aware Attention.} 
Then, to capture the legal-rational information of the legal case $X$ related to a specific law article $L_k$, we propose an article-aware attention mechanism. 
Our article-aware attention mechanism borrows from the self-attention mechanism of Transformer~\cite{vaswani2017Transformer} with three parts: query, key, and value.
The difference is that in LCM-LAI, we define a memory vector for each law article and the query of our attention mechanism is from these memory vectors instead of from the input sentence embeddings themselves.
That is, $\mathbf{q}_k= W_q\mathbf{m}_k$, $\mathbf{k}_i = W_k\mathbf{h}_i$, $\mathbf{v}_i = W_v\mathbf{h}_i$,
where $\textbf{m}_k \in \mathbb{R}^{d_b}$ is the memory vector associated to law article $L_k$ and
$W_k, W_q, W_v \in \mathbb{R}^{d_b \times d_h}$ are the trainable weighted parameters.
For simplicity, we denote the set of these memory vectors as $\mathcal{M} = \{ \mathbf{m}_1, \ldots, \mathbf{m}_{n_L}\}$.
Then, based on the query $\mathbf{q}_k$ and key $\mathbf{k}_i$, we compute the correlation score $\lambda_{i, k}$ between sentence $x_i$ and law article $L_k$, i.e.,
\begin{equation} \label{eq: sub_attention}
\lambda_{i, k} = \mathbf{q}_k^{\intercal} \mathbf{k}_i.
\end{equation}
Further, the output of our article-aware attention mechanism is calculated as
\begin{equation} \label{eq: sub_attention_sum}
\mathbf{X}_k = \sum_{i=1}^{n_x}\gamma_{i,k}\mathbf{v}_i,
\end{equation}
where $\gamma_{i, k} = \frac{\exp(\lambda_{i, k})}{\sum_{i=1}^{n_x} \exp(\lambda_{i, k})}$ denotes the normalized attention score,
$\mathbf{X}_k \in \mathbb{R}^{d_h}$ represents the legal-rational representation of legal case $X$ related to law article $L_k$.
For convenience narration, we call $\mathbf{X}_k$ as the $L_k$-rational representation of case $X$.
Intuitively, on the one hand, our attention mechanism captures the most relevant legal information for each law $L_k$ in case $X$, which is conducive to the article prediction sub-task.
On the other hand, we generate an interesting intermediate variable $\lambda_{i, k}$, which evaluates the legal-rational correlation of each sentence $x_i$ concerning the corresponding law $L_k$, which provides a basis for our subsequent calculation of the legal-rational correlation between sentences across cases from the perspective of each law article.

\noindent {\bf Article based Classifier.} 
Since the focus of our work is not to solve the multi-label classification, we simply treat the article prediction sub-task as multiple binary classifications.
To predict whether a law article is suitable for a case, we first compute the prototype of each law article.
Formally, we define the prototype of law article $L_k$ as $\mathbf{Proto}_k = \text{MLP}(\mathbf{L}_k) \in \mathbb{R}^{d_h}$.
Then, we use the cosine distance-based classifier to compute the probability that law article $L_k$ applies to case $X$, i.e.,
\begin{equation} \label{eq: article prediction}
\text{P}(L_k|X) = \sigma \left( \cos \left( \mathbf{X}_k, \mathbf{Proto}_k \right) \right)= \text{Sigmoid}\left(\frac{\mathbf{X}_k^\mathsf{T} \mathbf{Proto}_k}
{ |\mathbf{X}_k| \cdot |\mathbf{Proto}_k|}\right).
\end{equation}
In this way, we get the set of predicted law articles for the legal case $X$ by the following formula,
\[
\hat{\mathcal{L}}_X = \{ L_k | \text{P}(L_k|X) > 0.5, L_k \in \mathcal{L}\}.
\]

\subsubsection{Legal Correlation Matrix.} \label{subsubsec: law correlation matrix}
To measure the legal-rational correlation between across-case sentences, a straightforward way is directly computing the cosine similarity based on their corresponding value vectors as Eq.~\ref{eq: semantic correlation matrix} does.
However, such an absolute distance-based measure is still semantic and cannot comprehensively evaluate the legal-rational relevance from the perspective of multiple legal attributes.
Fortunately, our article-aware attention mechanism has computed the legal-rational correlation between sentences and law articles, i.e., 
$\lambda_{i, k}$ in Eq.~\ref{eq: sub_attention}.
Combining with the inspiration from Yurochkin et al~\cite{yurochkin2019topicdistribution}, who use the distribution of topic distribution to evaluate the relevance between documents, we consider evaluating the legal-rational correlation between across-case sentences from the perspective of law distribution.
Denoting the matrix with entries $\lambda_{i, k}$ as $\Lambda$,
the column vector $\lambda_{:, k}$ of $\Lambda$ can be used to capture the $L_k$-rational representation as it records which sentences are more related to the corresponding law article $L_k$ in legal-ration (cf. Eq.~\ref{eq: sub_attention}).
Meanwhile, we observe that the row vector $\lambda_{i, :}$ can also reflect the legal-rational correlation between the $i$-th sentence and all law articles, which can be regarded as the law distribution vector of the sentence, as depicted in Fig.~\ref{fig: Law Interaction}.
In what follows we simply write $\lambda_{i, :}$ as $\lambda_{i}$ when no confusion arises.
To compute the element $c_{i,j}^{(L)}$ of legal correlation matrix $\mathbf{C}^{(L)} \in \mathbb{R}^{n_x \times n_y}$, we compute the cosine similarity between the law distribution vectors of $x_i$ and $y_j$, that is,
\[
c_{i,j}^{(L)} = \cos(\mathrm{\lambda}_{i}^{X}, \mathbf{\lambda}_{j}^{Y}) = 
 \frac{{\mathbf{\lambda}_{i}^{X}}^\mathsf{T} \mathbf{\lambda}_{j}^{Y}}
 { |\mathbf{\lambda}_{i}^{X}| \cdot |\mathbf{\lambda}_{j}^{Y}|},
\]
where $\mathrm{\lambda}_{i}^{X}, \mathrm{\lambda}_{j}^{Y} \in \mathbb{R}^{n_L}$ separately denote the law distribution vectors of the corresponding sentence $x_i$ in $\Lambda^{X} $ and $y_j$ in $\Lambda^{Y}$.
Such a distribution-based similarity can effectively measure the law-rational correlation because the high-similarity sentence pair must have almost similar legal attributes (i.e., similar relation to all law articles) instead of simply having a small distance in the semantic space.

\subsubsection{Article-intervened Interaction Encoder.}\label{subsubsec: law interaction encoder}
To extract the legal-rational matching information between pairwise legal cases, according to the process of Sec.~ \ref{subsubsec: sentence interaction}, we first get the weight matrices of $\alpha_{i, j}^{(L)}$ and $\beta_{i, j}^{(L)}$ referring to the Eq.~\ref{eq: Semantic Interaction Attention}.
As the value vectors in the article-aware attention module (i.e., $\textbf{v}_i$ in Eq.~\ref{eq: sub_attention_sum}) are closely associated with the article prediction sub-task, we leverage them to obtain the legal interaction representation of each sentence (cf. Fig.~\ref{fig: Law Interaction}), that is,
\begin{equation} \label{eq: legal_inter_rep}
\mathbf{x}_i^{(L)} = \mathbf{v}_i^X \oplus \sum_{j=1}^{n_y}\alpha_{i, j}^{(L)}\mathbf{v}_j^Y;
\ \ 
\mathbf{y}_j^{(L)} = \mathbf{v}_j^Y \oplus \sum_{i=1}^{n_x}\beta_{i, j}^{(L)}\mathbf{v}_i^X,
\end{equation}
where $\mathbf{v}_i^X, \mathbf{v}_j^Y \in \mathbb{R}^{d_h}$ denote the value vectors of the corresponding sentences $x_i$ and $y_j$, 
and $\mathbf{x}_i^{(L)}, \mathbf{y}_j^{(L)} \in \mathbb{R}^{2d_h}$ denote the corresponding legal interaction representations.
Following similar calculation of Eqs.~\ref{eq: Semantic RNN} and~\ref{eq: Semantic Representation}, we get the legal hidden states (i.e., $\textbf{H}_{X}^{(L)} = [\mathbf{h}_{x,1}^{(L)}, \cdots, \mathbf{h}_{x,n_x}^{(L)}]$ and $\textbf{H}_{Y}^{(L)} = [\mathbf{h}_{y,1}^{(L)}, \cdots, \mathbf{h}_{y,n_y}^{(L)}]$) and the legal interaction representations (i.e., $\mathbf{X}^{(L)}$ and $\mathbf{Y}^{(L)}$) for pairwise legal cases.
We use $d_l$ to represent the dimension of the legal interaction representations, i.e., $\mathbf{X}^{(L)}, \mathbf{Y}^{(L)} \in \mathbb{R}^{d_l}$.

Besides, to avoid the interference caused by similar background statements (i.e., sentences with very uniform law distribution), we propose an \textbf{a}rticle-\textbf{i}ntervened \textbf{a}ttention (AIA) mechanism to focus on the interaction representation of legal-rational prominent sentences. 
Take the specific legal case $X$ as an example, where we first use a mean operation to aggregate the predicted applicable law article to a context vector, i.e.,
\[
\Phi_{X} = \frac{1}{|\hat{\mathcal{L}}_{X}|} \sum_{L_k\in \hat{\mathcal{L}}_X} \mathbf{L}_k,
\]
where $\Phi_{X} \in \mathbb{R}^{d_b}$ denotes the context vector and $|\hat{\mathcal{L}}_{X}|$ is the number of predicted law article set in the article prediction subtask.
Then, we first calculate the legal-ration-related attention score of each sentence according to the context vector, that is,
\[
\psi_i = \frac{\exp\left((\mathbf{W}_h\mathbf{h}_{x,i}^{(L)})^\mathsf{T}(\mathbf{W}_{\Phi}\Phi_{X})\right)}{\sum_{i=1}^{n_x}\exp\left((\mathbf{W}_h\mathbf{h}_{x,i}^{(L)})^\mathsf{T}(\mathbf{W}_{\Phi}\Phi_{X})\right)},
\]
where $\mathbf{W}_h \in \mathbb{R}^{d_h \times d_l} $ and $\mathbf{W}_{\Phi} \in \mathbb{R}^{d_h \times d_b}$ denote the trainable parameter matrices.
Thus, we attentively aggregate the hidden states into a novel representation, i.e.,
\[
\mathbf{X}^{(A)} = \sum_{i=1}^{n_x}\psi_i\mathbf{h}_{i}^{(L)},
\]
where $\mathbf{X}^{(A)} \in \mathbb{R}^{d_l}$ and we call it the article-intervened representation of case $X$.

\subsection{Matching Prediction}
When we predict the final matching results, we concatenate the semantic interaction representation, legal interaction representation, and article-intervened representation as the final representation, that is,
\[
\mathbf{X}_f = \mathbf{X}^{(S)} \oplus \mathbf{X}^{(L)} \oplus \mathbf{X}^{(A)};
\ \ 
\mathbf{Y}_f = \mathbf{Y}^{(S)} \oplus \mathbf{Y}^{(L)} \oplus \mathbf{Y}^{(A)},
\]
where $\mathbf{X}_f, \mathbf{Y}_f \in \mathbb{R}^{d_s + 2d_l}$ denote the final representations of legal cases $X$ and $Y$.

For the LCM task treated as a classification task, we adopt the classification object function of Sentence-BERT~\cite{reimers2019sentence-BERT} to obtain matching results.
Specifically, we concatenate the case embedding $\mathbf{X}_f$ and $\mathbf{Y}_f$ with their element-wise difference $|\mathbf{X}_f - \mathbf{Y}_f|$ and put them into a linear classifier:
\[ 
\hat{\mathbf{Z}}(X, Y) = \text{softmax}\left(\mathbf{W}_p\left(\mathbf{X}_f \oplus \mathbf{Y}_f \oplus |\mathbf{X}_f - \mathbf{Y}_f|\right)\right),
\]
where $\mathbf{W}_p \in \mathbb{R}^{3(d_s + 2d_l) \times |Z|}$ is a trainable parameter matrix, $\hat{\textbf{Z}}(X, Y) \in \mathbb{R}^{|Z|}$ is the predicted matching vector of the legal case pair $(X, Y)$, and $|Z|$ is the number of matching labels.

As the LCR task, we treat it as a rank task and use the cosine distance to measure the matching score of pairwise legal cases:
\[ 
\hat{S}(X, Y) = \frac{\mathbf{X}_f^\mathsf{T} \mathbf{Y}_f}
{|\mathbf{X}_f| \cdot |\mathbf{Y}_f|}.
\]

\subsection{Loss Function}
We treat the article prediction, LCM, and LCR tasks as different basic deep learning tasks, which also face some additional challenges for model training.
In this subsection, we introduce our loss function selection and why we choose them.

\noindent {\bf Article Prediction Loss.}
When treating the article prediction sub-task as a multi-label classification, it faces the imbalance problem that the number of negative samples is far greater than that of positive ones for most law articles, since common charges constitute the vast majority of legal cases.
Such a challenge leads to the non-convergence of this sub-task when we apply the traditional binary cross-entropy loss function for training, which further affects the performances of final LCR \& LCM tasks.
Since the existing works~\cite{cai2024ner,tong2024legal, chen2024knowledge} usually adopt the ZLPR loss proposed by Su et al.~\cite{su2022ZLPR_loss} as the objective function when solving similar challenge tasks, we choose it as our loss function.
Take the specific case $X$ as an example, the ZLPR loss first computes the matching vector $\mathbf{S}_{X} \in \mathbb{R}^{n_L}$ about all classes, that is,
\[
\mathbf{S}_{X} = \tau_{a} \cdot \left[\cos(\mathbf{X}_1, \mathbf{Proto}_1), \ldots, \cos(\mathbf{X}_{n_L}, \mathbf{Proto}_{n_L})\right],
\]
where $\tau_{a}$ is a temperature coefficient. Then, according to the article ground-truth vector $\mathbf{Z}_{L}^{X} \in \mathbb{R}^{n_L}$, the ZLPR loss uses a rank-based function to constrain the score of the positive label to be higher than the score of the negative label. The specific formula is as follows,
\[
\mathscr{L}_{a}^{X} = \log\left(1 + \mathbf{Z}_{L}^{X} \odot e^{-\mathbf{S}_{X}}\right) + \log\left(1 + \left(1-\mathbf{Z}_{L}^{X}\right) \odot e^{\mathbf{S}_{X}}\right),
\]
where $\odot$ denotes the inner product operator.
Different from the BCE is only focused on the binary relevance of every single label, 
the ZLPR loss additionally considers the correlation between labels.
This makes it more comprehensive and robust than the former.

\noindent {\bf Loss for LCR.} 
When we solve the LCR task as a rank task, to fully use the fine-grained match-level labels,
referring to the existing works~\cite{gong2023transferable,liu2022ynu},
we choose the CoSENT loss~\cite{huang2024cosent} as the objective function to optimize the cosine similarity between legal cases.
The main idea of CoSENT loss is to constrain that the samples with the higher-level label must get higher scores than those with a lower-level label. Here, we give the specific formula,
\[
\mathscr{L}_{m} = \log\left(1+ \sum_{\text{sim}(X_{i}, Y_{j}) \textgreater \text{sim}(X_{m}, Y_{n})} e^{\tau_{m}(\hat{S}(X_{i}, Y_{j})- \hat{S}(X_{m}, Y_{n}))}\right),
\]
where $\tau_{m}$ is a temperature coefficient.
Note that $\text{sim}(X_{i}, Y_{j}) \textgreater \text{sim}(X_{m}, Y_{n})$ means the legal case pair $(X_i, Y_j)$ has the higher match-level than the legal case pair $(X_m,Y_n)$.
Thus, compared with the typical triplet loss~\cite{chechik2010TripletLoss} whose contrastive form targets an anchor sample, CoSENT loss has a more general contrastive form.

\noindent {\bf Loss for LCM.} As for the LCM task, we take it as a typical classification task and compute the cross-entropy (CE) loss as its objective function, that is,
\[
\mathscr{L}_{m} = -\sum_{i=1}^{|Z|} {z}_{i}\log({\hat{z}_{i}}),
\]
where $\hat{z}_{i}$ is the $i$-th element of the predict vector $\hat{\textbf{Z}}(X, Y)$ and $\textbf{Z}(X, Y) = [z_1, \ldots, z_{|Z|}]$ denotes the one-hot ground-truth matching vector of legal case pair $(X, Y)$.

In summary, the overall loss for full training is the sum of the subtask and the main task, that is,
\[
\mathscr{L} = \mathscr{L}_{a} + \mathscr{L}_{m}.
\]

\section{Experiments} \label{sec:experiments}
In this section, we conduct experiments to demonstrate the effectiveness of LCM-LAI by answering the following research questions:

\noindent {\bf RQ1:} Can LCM-LAI outperform the existing state-of-the-art
baselines on LCR \& LCM tasks?

\noindent {\bf RQ2:} Is jurisprudential relevance essential for legal case retrieval and matching? Does the dependent multi-task learning framework with the law article prediction subtask improve the performance of LCM-LAI?

\noindent {\bf RQ3:} Can LCM-LAI meet the efficiency limitations of practical applications?

\noindent {\bf RQ4:} Whether LCM-LAI is explicable for the LCR \& LCM prediction results?

\begin{table}[t]
\centering
\caption{ Statistics of LeCaRD, LeCaRDv2, ELAM, and eCAIL datasets.}\label{tab: Data Statistics}
\begin{tabular}{l|c|c|c|c}
\toprule
     & LeCaRd   & LeCaRDv2     & ELAM  & eCAIL\\ 
 \midrule
 Suitable for LCR           & \checkmark    & \checkmark    & $\times$  & $\times$ \\
 \# Query                   & $107$         & $800$         & $--$      & $--$ \\
 \# Candidates per query    & $30$         & $30$      & $--$      & $--$ \\
 \# Train Query             & $85$          & $640$         & $--$      & $--$ \\
 \# Valid Query             & $11$          & $80$          & $--$      & $--$ \\
 \# Test Query              & $11$          & $80$         & $--$      & $--$ \\
 \midrule
 Suitable for LCM           & \checkmark    & $\times$  & \checkmark    & \checkmark \\
 \# Case pairs              & $3,210$       & $--$      & $5,000$       & $7,500$ \\
 \# Label levels            & $4$           & $--$      & $3$           & $3$ \\
 \# Train pairs             & $2,582$       & $--$      & $4,000$       & $6,000$ \\
 \# Valid pairs             & $323$         & $--$      & $500$         & $750$ \\
 \# Test pairs              & $323$         & $--$      & $500$         & $750$ \\
 \midrule
 \# Law articles            & $54$          & $86$      & $42$          & $ 107 $ \\
 Avg. \# cited law articles per case     & $ 12.74 $   & $4.84$   & $5.04$   & $4.59$ \\
 Avg. \# sentences per case     & $97.86$   & $16.42$   & $16.29$   & $124.10$ \\
 \bottomrule
\end{tabular}
\end{table}
\subsection{Experimental Settings}
\subsubsection{Datasets.} We validate our model on experiments of both LCR and LCM tasks. Here we introduce the four available datasets we used:
LeCaRD~\cite{ma2021LeCaRD}, LeCaRDv2, ELAM~\cite{yu2022Explainable}, and eCAIL~\cite{yu2022Explainable}.

\textbf{LeCaRD}\footnote{\url{https://github.com/myx666/LeCaRD}} is a legal case retrieval dataset that contains 107 query cases and 100 candidate cases per query selected from more than 43,000 criminal legal cases, which are published by the Supreme People’s Court of China.
As the top 30 candidate cases for each query are manually annotated with a $4$-level relevance (matching) label, this dataset is used for both LCR and LCM tasks.

\textbf{LeCaRDv2}\footnote{\url{https://github.com/THUIR/LeCaRDv2}} is an extension of the LeCaRD dataset.
Compared with LeCaRD, the data size of LeCaRDv2 is approximately $8$ times larger. 
Specifically, the LeCaRDv2 dataset comprises 800 query cases and 55,192 candidate cases extracted from 4.3 million criminal case documents. 
It gives a candidate set with $30$ legal cases for each query, which is suitable for evaluation on the LCR task.

\textbf{ELAM}\footnote{\url{https://github.com/ruc-wjyu/IOT-Match}} is annotated to solve the explainable legal case match task.
For each pair of legal cases, a $3$-level manually assigned matching label is given: $0$, $1$, and $2$ refer to a mismatch, a partial match, and a complete match, respectively.
In addition to the matching labels, fine-grain explainable labels are also provided in this dataset, such as the rationale labels of sentences, the alignments of rationales, and the literal explanations.
In the experiment, we only use the fact description of legal cases and the matching labels for training and evaluation.

\textbf{eCAIL} is extended from the CAIL 2021 dataset\footnote{The open sourced data for the Fact Prediction Track: \url{http://cail.cipsc.org.cn/}}, following the practice in~\cite{yu2022Explainable}. 
Each legal case is annotated in the original dataset with some tags about private lending.
Each case pair of the eCAIL dataset gets a $3$-level matching label based on the number of overlapping tags: overlapping $\textgreater 10$ and overlapping $\textless 1$ refer to matching and mismatching respectively, while other ranges indicate partial matching.
Similar to the ELAM dataset, we also only use matching labels for the training of LCM-LAI.

As for the selection of law articles, referring to previous works on the JLP task, we only keep the law articles applied to not less than $10$ corresponding case samples in the dataset.
Notice that, for LeCaRD, LeCaRDv2, and ELAM, we use the PRC Criminal Law.
For eCAIL, we use the PRC Contract Law\footnote{The contents of law articles can be downloaded from: \url{http://flk.npc.gov.cn/}.}.
The detailed statistics of these datasets are shown in Tab.~\ref{tab: Data Statistics}.

\subsubsection{Baselines.}
We compare our LCM-LAI with the following LCR \& LCM baselines, including the universal models (with the \textbf{[R\&M]} tag), LCR-specific models (with the \textbf{[R]} tag), and LCM-specific models (with the \textbf{[M]} tag).

\noindent{\bf [R\&M] Sentence-BERT~\cite{reimers2019sentence-BERT}:} 
a BERT-based text-matching model that uses BERT~\cite{devlin2018bert} to encode two input cases separately and then uses an MLP to conduct matching by inputting the concatenation of two case embeddings.

\noindent{\bf [R\&M] Lawformer~\cite{xiao2021Lawformer}:}
a Longformer-based pre-trained language model designed to enhance the representation of lengthy legal documents. 
This model has been trained using millions of Chinese legal cases.
In our experiment, we send the concatenation of two cases to Lawformer and use the mean pooling of output to conduct matching.

\noindent{\bf [R\&M] BERT-PLI~\cite{shao2020Bert_PLI}:}
a text-matching model that first uses BERT to capture paragraph-level semantic relationships of two documents and leverages an RNN with an attention mechanism to aggregate the relation embeddings to calculate the matching score by a binary classifier.

\noindent{\bf [R\&M] Thematic Similarity \cite{bhattacharya2020LCR_survey}:} another text-matching model considering the fine-grain similarity that segments two legal cases into paragraphs and computes the paragraph-level similarity. Unlike BERT-PLI, this method uses maximum or average similarities for matching prediction. Combined with the CoSENT loss function, we also compare LCM-LAI to it on the LCR task.

\noindent{\bf [R\&M] ColBERT \cite{khattab2020colbert}:} a classical multi-representation dense search method with a late interaction framework.
Different from Thematic Similarity, which learns only one embedding for each sentence or paragraph, this method learns embeddings for each token and then uses a “MaxSim” operation to compute the matching score between sentences across the query and candidates.

\noindent{\bf [R\&M] ColBERT-X\footnote{In this paper, we load the checkpoint in a bilingual Chinese and English version: \url{https://huggingface.co/hltcoe/plaidx-large-zho-tdist-mt5xxl-engzho}} \cite{ecir2022colbert-x}}: a pre-trained dense retrieval mode with the ColBERT framework for the cross-language information retrieval (CLIR) task.

\noindent{\bf [M] Law-Match~\cite{sun2023law}:} a model-agnostic method that uses the corresponding cited law article to reconstruct the representation of legal cases.
In the setting of the inference stage of our LCR task, the query is original without labels for applicable law articles, which makes Law-Match unfit for the LCR task. 
On the LCM task, our experiment only reports the best performance of the three variants that connect the Law-Match module with Sentence-BERT, Lawformer, or BERT-PLI backbone.

\noindent{\bf [M] IOT-Match~\cite{yu2022Explainable}:} an explainable legal case matching model that additionally marks the rationale labels of case sentences and the alignments labels of rationales with a significant cost. Then, it leverages these elaborate labels to not only improve the performance of the LCM task but also generate natural language-based explanations for the predicted matching results. 
Due to the lack of rationale labels in the LeCaRD and LeCaRDv2 datasets, this method only suits the ELAM and eCAIL datasets for the LCM task. 

\noindent{\bf [R] Some BOW retrieval models.} The typical models include TF-IDF, BM25~\cite{robertson1994BM25}, and LMIR~\cite{song1999LMIR}.

\subsubsection{Evaluation metrics.} For different tasks, we introduce the selected suitable labels for performance evaluation here.

For the LCR task, we employ the top-$k$ Normalized Discounted Cumulative Gain (N@$k$), Precision (P@$k$), and Mean Average Precision (MAP) as evaluation metrics. Among them, we mainly evaluate the P@$k$ and MAP metrics because the LCR task focuses on the candidate cases that perfectly match the query case.

For the LCM task, we choose four typical metrics that are widely used for classification tasks, including accuracy (Acc.), macro-precision (MP), macro-recall (MR), and macro-F1 score. 
Among them, we mainly evaluate all models with the F1 score, which more objectively reflects the effectiveness of our LCM-LAI and other baselines. 
Besides, we also use a Mean-F1 score to evaluate the overall performance, which is the mean of the F1 score on all three datasets.

\subsubsection{Implementation details.} \label{sec: hyperparametter_setting}
For input of the LCM-LAI, we set the maximum case length as 15 sentences and the maximum sentence length as 150 tokens.
As for the hyper-parameters, we use grid search on the validation set of the LCM tasks with AdamW~\cite{loshchilov2018AdamW} optimizer to obtain the best set of LCM-LAI.
The dimension of interaction-related representations (i.e., $d_s$ and $d_l$) is set as $2 \times 768$ as the dimension of all other latent states (i.e., $d_b$ and $d_h$) is set as $768$.
The temperature coefficient $\tau_{a}$ and $\tau_{m}$ respectively are $10$ and $20$.
For training, we set the learning rate of the AdamW optimizer to $3e-5$ and the batch size to $8$.
We set the hyper-parameters as the optimal values in the original paper for all baselines.
In either the LCR task or the LCM task, we train each model for 50 epochs and choose the model with the best score on our focused metric (i.e., MAP and F1 score) on the validation set for testing.
To ensure the reliability of the experimental results, we take the $5$-fold settings for all tasks.
Specifically, as referenced in Tab.~\ref{tab: Data Statistics}, we have partitioned all datasets to ensure that the training set occupies 4/5 of the total, with the remaining 1/5 being evenly divided into the validation set and test set.
Thus, in our experiment, we repeat the experiment five times and make sure the 1/5 data used for validation and testing in each experiment is non-overlapping.
Finally, for each metric, we report the mean of five experiments for comparison.
In addition, for all models that take BERT as the backbone (i.e., our LCM-LAI and all baselines except Lawformer, ColBERT-X, and BOW retrieval models), we initialize them with the parameters of the Legal-Bert\footnote{\url{https://github.com/thunlp/OpenCLaP}}.
Besides, for the ColBERT and ColBERT-X, we split the case into paragraphs and use the MaxP trick (taking the maximum passage score among a document as the document score) to make them adaptive to the document length in LCR \& LCM tasks.
\begin{table*}[t]
\centering
\caption{
The retrieval performances on LeCaRD and LeCaRDv2 test sets.
The best results are in \textbf{bold} and the second ones are \underline{underlined}.
$\dagger$ denotes that LCM-LAI achieves significant improvements over all baselines in paired t-test with $p$-value $\textless 0.05$.
}
\label{tab: Retrieval Performance}
\resizebox{\linewidth}{!}{%
\begin{tabular}{lcccccccccccc}
\toprule
 Datasets            & \multicolumn{6}{c}{LeCaRD}        & \multicolumn{6}{c}{LeCaRDv2}\\ 
\cmidrule(lr){2-7} \cmidrule(lr){8-13} 
 Metrics        & N@10       & N@20       & N@30       & MAP       & P@10      & P@5
                & N@10       & N@20       & N@30       & MAP       & P@10      & P@5\\
 \midrule
 BM25               & $ 74.27 $     & $ 80.01 $     & $ 88.98 $     & $ 46.18 $     & $ 39.50 $      & $ 37.57 $ 
                    & $ 82.97 $     & $ 86.84 $     & $ 93.16 $     & $ 84.35 $     & $ 82.25 $      & $ 82.00 $     \\
                    
 TF-IDF             & $ 68.09 $     & $ 76.21 $     & $ 86.34 $     & $ 41.04 $     & $ 33.46 $      & $ 32.57 $ 
                    & $ 80.84 $     & $ 85.03 $     & $ 92.32 $     & $ 82.84 $     & $ 79.25 $      & $ 78.75 $     \\

 LMIR               & $ 75.52 $     & $ 80.50 $     & $ 88.50 $     & $ 47.74 $     & $ 39.69 $      & $ 43.11 $ 
                    & $ 83.06 $     & $ 86.62 $     & $ 93.05 $     & $ 84.70 $     & $ 84.00 $      & $ 83.25 $     \\
\midrule
 BERT               & $ 77.41 $     & $ 81.80 $     & $ 89.36 $     & $ 51.67 $     & $ 41.76 $      & $ 45.25 $ 
                    & $ 86.99 $     & $ 89.98 $     & $ 94.53 $     & $ 89.38 $     & $ 88.38 $      & $ 89.75 $     \\
                    
 sentence-BERT      & $ 79.25 $     & $ 83.66 $     & $ 89.89 $     & $ 50.93 $     & $ 43.20 $      & $ 46.71 $ 
                    & $ 86.66 $     & $ 89.82 $     & $ 94.36 $     & $ 89.44 $     & $ 89.00 $      & $ 89.75 $     \\
            
 
 BERT-PLI           & $ 79.33 $     & $ 84.06 $     & $ 90.23 $     & $ 53.64 $     & $ 44.47 $      & $ 47.50 $ 
                    & $ 87.83 $     & $ 90.91 $     & $ 94.83 $     & $ 89.93 $     & $ 88.87 $      & $ 90.50 $     \\
                    
 Lawformer          & $ 78.30 $     & $ 84.06 $     & $ 90.24 $     & $ 52.84 $     & $ 43.40 $      & $ 46.77 $ 
                    & $ \underline{89.15} $     & $ 90.47 $     & $ 95.36 $     & $ 89.53 $     & $ \underline{89.75} $      & $ 90.50 $     \\

 Thematic Similarity(avg) 
                    & $ 79.88 $     & $ 84.13 $     & $ 90.45 $     & $ 51.22 $     & $ 41.13 $      & $ 44.45 $ 
                    & $ 89.00 $     & $ \underline{91.58} $     & $ \underline{95.53} $     & $ 90.25 $     & $ 89.00 $      & $ 90.50 $     \\

 Thematic Similarity(max) 
                    & $ 79.27 $     & $ 83.71 $     & $ 89.97 $     & $ 51.73 $     & $ 43.23 $      & $ 47.21 $ 
                    & $ 87.35 $     & $ 90.21 $     & $ 94.42 $     & $ 88.83 $     & $ 87.87 $      & $ 89.25 $     \\

 ColBERT 
                    & $ 79.89 $     & $ 84.25 $     & $ 90.38 $     & $ 53.90 $     & $ 42.68 $      & $ 46.77 $ 
                    & $ 87.49 $     & $ 90.52 $     & $ 94.76 $     & $ 89.38 $     & $ 89.12 $      & $ 89.75 $     \\

 ColBERT-X 
                    & $ \underline{80.41} $     & $ \underline{85.21} $     & $ \underline{90.65} $     & $ 53.51 $     & $ 44.00 $      & $ 47.17 $ 
                    & $ 87.51 $     & $ 90.63 $     & $ 94.65 $     & $ 90.19 $     & $ 89.13 $      & $ 90.75 $     \\
 
 \midrule

 \bf{LCM-LAI + CE}       
                & $ 79.93 $     & $ 83.69 $     & $ 89.87 $     & $ \underline {54.42^{\dagger}} $     & $ \underline {44.54} $      & $ \underline {48.13}^{\dagger} $ 
                & $ 89.01 $     & $ 91.20 $     & $ 95.17 $     & $ \underline{90.79}^{\dagger} $     & $ 89.38 $      & $ \bf{92.50}^{\dagger} $     \\

 \bf{LCM-LAI + CoSENT}       
                & $ \bf {81.33}^{\dagger} $     & $ \bf {85.72}^{\dagger} $     & $ \bf {91.18}^{\dagger} $     & $ \bf {56.69}^{\dagger} $     & $ \bf {45.00}^{\dagger} $      & $ \bf {49.38}^{\dagger} $ 
                & $ \bf{89.63}^{\dagger} $     & $ \bf{91.90}^{\dagger} $     & $ \bf{95.70} $     & $ \bf{91.27}^{\dagger} $     & $ \bf{90.38}^{\dagger} $      & $ \underline{91.75}^{\dagger} $     \\
 \bottomrule
\end{tabular}%
}
\end{table*}

\subsection{Basic Performance Evaluation (RQ1)}
\subsubsection{The LCR Results.} 
We present the results of our experiments conducted on the LeCaRD and LeCaRDv2 datasets for the LCR task, as shown in Tab.~\ref{tab: Retrieval Performance}.
In summary, LCM-LAI demonstrates exceptional performance across all six metrics compared to baseline methods, showcasing its state-of-the-art capabilities.
On the LeCaRD dataset, LCM-LAI showcases significant enhancements in MAP, P@5, and P@10, with improvements of $2.79\%$, $1.88\%$ and $0.53\%$ respectively, surpassing the state-of-the-art baseline. Similarly, the LeCaRDv2 dataset also witnessed improvements of $1.02\%$, $1.75\%$, and $0.63\%$.

\begin{table*}[t]
\centering
\caption{
The matching performances on ELAM, LeCaRD, and eCAIL test sets.
The best results are in \textbf{bold} as the second ones are \underline{underlined}.
$\dagger$ denotes that LCM-LAI achieves significant improvements over all baselines in paired t-test with $p$-value $\textless 0.05$.
“$--$” denotes that the IOT-match fails to deal with the ELAM dataset due to a lack of necessary input or label.
Thus, we use the average F1 score of other methods on the LeCaRD dataset to compute its final Mean-F1 score.
}\label{tab: Match Performance}
\resizebox{\linewidth}{!}{%
\begin{tabular}{lccccccccccccc}
\toprule
 Datasets              & \multicolumn{4}{c}{ELAM}                     & \multicolumn{4}{c}{LeCaRD}   & \multicolumn{4}{c}{eCAIL} & \multirow{2}{*}{Mean-F1}\\ 
 \cmidrule(lr){2-5} \cmidrule(lr){6-9} \cmidrule(lr){10-13} 
 Metrics            & Acc.        & MP          & MR          & F1      
                    & Acc.        & MP          & MR          & F1      
                    & Acc.        & MP          & MR          & F1        \\
 \midrule

 BERT                       & $ 68.58 $     & $ 68.46 $     & $ 67.86 $     & $ 67.97 $     
                            & $ 62.35 $     & $ 64.49 $     & $ 61.19 $     & $ 62.11 $     
                            & $ 73.68 $     & $ 74.07 $     & $ 73.71 $     & $ 73.76 $     & $67.95$\\
                            
 Sentence-BERT              & $ 72.47 $     & $ 71.50 $     & $ 71.47 $     & $ 71.29 $     
                            & $ 62.29 $     & $ 61.34 $     & $ 60.40 $     & $ 60.59 $     
                            & $ 77.87 $     & $ 78.18 $     & $ 77.92 $     & $ 77.99 $     & $69.96$\\
                            
 BERT-PLI                   & $ 73.59 $     & $ 73.43 $     & $ 72.88 $     & $ 73.08 $     
                            & $ 66.44 $     & $ 68.80 $     & $ 65.21 $     & $ 66.57 $     
                            & $ 77.87 $     & $ 77.82 $     & $ 78.06 $     & $ 77.83 $     & $72.49$\\
                    
 Lawformer                  & $ 73.47 $     & $ 73.30 $     & $ 72.55 $     & $ 72.70 $     
                            & $ 66.08 $     & $ 66.41 $     & $ 64.43 $     & $ 65.45 $     
                            & $ 77.98 $     & $ 78.16 $     & $ 78.03 $     & $ 78.04 $     & $72.06$ \\

 Thematic Similarity(avg)   & $ 73.79 $     & $ 71.92 $     & $ 72.30 $     & $ 71.93 $     
                            & $ 62.26 $     & $ 64.61 $     & $ 60.74 $     & $ 61.82 $        
                            & $ 76.00 $     & $ 75.68 $     & $ 76.06 $     & $ 75.69 $     & $69.81$\\

 Thematic Similarity(max)   & $ 73.71 $     & $ 72.80 $     & $ 72.81 $     & $ 72.57 $     
                            & $ 65.08 $     & $ 67.57 $     & $ 62.62 $     & $ 63.81 $     
                            & $ 77.44 $     & $ 77.34 $     & $ 77.50 $     & $ 77.09 $     & $71.16$\\
 ColBERT   & $ 74.27 $     & $ 73.29 $     & $ 73.31 $     & $ \underline{73.19} $ 
                            & $ 67.12 $     & $ 68.26 $     & $ 66.34 $     & $ 66.85 $  
                            & $ 78.24 $     & $ 78.43 $     & $ 78.37 $     & $ 78.35 $  & $ 72.80 $  \\

 ColBERT-X & $ 74.39 $     & $ 72.91 $     & $ 72.71 $     & $ 72.50 $     
                            & $ 67.68 $     & $ 68.92 $     & $ 64.53 $     & $ 65.78 $    
                            & $ 78.61 $     & $ 78.43 $     & $ 78.76 $     & $ 78.66 $     & $ 72.31 $  \\

 IOT-Match                  & $ 73.87 $     & $ 73.02 $     & $ 72.41 $     & $ 72.55 $     
                            & $ -- $     & $ -- $     & $ -- $     & $ -- $     
                            & $ \bf 82.00 $     & $ \bf 82.10 $     & $ \bf 81.92 $     & $ \bf 81.90 $     & $73.09$\\

 Law-Match                  & $ \underline{74.95} $     & $ 72.96 $     & $ 71.75 $     & $ 72.35 $     
                            & $ 65.63 $     & $ 66.07 $     & $ 63.75 $     & $ 64.41 $     
                            & $ 80.00 $     & $ 79.78 $     & $ 79.92 $     & $ 79.84 $     & $72.20$\\
                            
\midrule
 \bf{LCM-LAI + CE}          
            & $\bf{76.51}^{\dagger} $ & $\bf{75.88}^{\dagger} $ & $\bf{75.64}^{\dagger} $ & $\bf{75.57}^{\dagger} $    
            & $\bf{70.97}^{\dagger} $ & $\bf{71.95}^{\dagger} $ & $\bf{71.36}^{\dagger} $ & $\bf{71.24}^{\dagger} $    
            & $\underline{81.07} $ & $\underline{81.01} $ & $\underline{81.23} $ & $\underline{80.89}$      & $\bf 75.90^{\dagger}$\\
 \bf{LCM-LAI + CoSENT}      
 & $ 74.31 $ & $\underline{73.61}^{\dagger} $ & $\underline{73.32}^{\dagger} $ & $ 73.15 $    
 & $\underline{69.10}^{\dagger} $ & $\underline{71.37}^{\dagger} $ & $\underline{69.02}^{\dagger} $ & $\underline{69.47}$ 
 & $ 80.08 $ & $ 80.16 $ & $ 80.04 $ & $ 79.91 $ & $\underline{74.18}^{\dagger}$\\

 \bottomrule
\end{tabular}%
}
\end{table*}
\subsubsection{The LCM Results.}
Tab.~\ref{tab: Match Performance} shows the experiment results of the LCM task on the ELAM, LeCaRD, and eCAIL datasets. 
Overall, our proposed LCM-LAI model exhibits the most competitive performance, which achieves the best Mean-F1 score of the three datasets with an improvement of $2.81\%$. 
Specifically, on the two datasets of all three, i.e., ELAM and LeCaRD, our LCM-LAI achieves state-of-the-art performance in all four evaluation metrics.
Compared with the state-of-the-art baseline, it improves the F1 scores by $2.49\%$ on dataset ELAM and $4.67\%$ on dataset LeCaRD.
As for the eCAIL dataset, LCM-LAI also gets the second-best performance, coming closest to the state-of-the-art IOT-Match model, with only a $1.01\%$ lower F1 score.

\subsubsection{Discussion of Vertical Association.} \label{sec: basic_performance}
We correlated the results of LCR \& LCM tasks longitudinally while comparing the baselines in each task horizontally.
Then we get some confirmatory observations and guesses as follows:
\begin{enumerate}
    \item {
    For the LCR \& LCM tasks, LCM-LAI outperforms the most related baseline Law-Match on all datasets across all metrics, even without the utilization of ground-truth applicable law articles as input.
    This is attributed to LCM-LAI's end-to-end dependent multi-task learning framework, which combines case legal-rational feature extraction with representation computation to learn more comprehensive representations and better performance.
    }
    \item {
    In addition, we focus on comparing our approach to the IOT-Match baseline, which heavily relies on numerous elaborate labels.
    These labels include legal-rational labels for each sentence and alignment labels across case sentences for effectively capturing interaction information through supervised learning in the LCM task.
    In the experiments on all four datasets, IOT-Match cannot be applied to LeCaRD and LeCaRDv2 datasets due to the lack of rationale annotations and sentence alignment annotations.
    Besides, it performs worse than our LCM-LAI on the ELAM dataset.
    This demonstrates the effectiveness and robustness (or generalization) of our LCM-LAI, as it performs well without elaborate expert annotation.
    It is noted that the performance of LCM-LAI on the e-CAIl dataset is inferior to IOT-Match.
    We argue it is because many legal definitions of civil cases are relatively more fine-grained and more complex, 
    in this situation, the additional careful annotation that IOT-Match used can provide more detailed domain knowledge than directly using legal text definitions.
    Despite bringing performance improvement, we must emphasize that these additional annotations also hinder the generalization ability of IOT-MATch under different legal systems.
    }
    \item {
    Comparing the performance between \textit{LCM-LAI + CL} and \textit{LCM-LAI + CoSENT}, as outlined in Tabs.~\ref{tab: Retrieval Performance} and~\ref{tab: Match Performance}, we observed that CoSENT loss outperforms CE loss in the LCR task, especially on the metrics of NDCG@$k$.
    This phenomenon can be attributed to the effective utilization of fine-grained match-level labels for learning ranking strategies, facilitated by the contrast term $\hat{S}(X_{i}, Y_{j})- \hat{S}(X_{m}, Y_{n})$ within the CoSENT loss.
    As a result, this mechanism aids in addressing the challenges associated with LCR tasks.
    Conversely, an opposite trend was observed in the LCM task, suggesting that a nonlinear classification surface is better suited for this classification-like task.
    }
    \item {
    Comparing multi-vector retrieval baselines with single-vector ones, we could find multi-vector based baselines are more advantageous. For example, even though ColBERT-X has not been pre-trained on legal data, it still achieves better overall performance than the single-vector ones like BERT, Sentence-BERT, BERT-PLI, Lawformer and Thematic Similarity (avg/max).
    However, their performance is still inferior to our LCM-LAI, further reflecting the importance and necessity of considering legal relevance between cases, which we emphasize in this paper.
    }
\end{enumerate}

\subsection{Ablation Experiments (RQ2)}
\subsubsection{\textbf{Ablation of the Framework}}
To further emphasize the importance of incorporating both semantic and jurisprudential relevance in legal cases, we conducted a series of ablation experiments on the fifth data division of the LeCaRD dataset on the LCM task to illustrate this point.

In this experiment, the specific ablation variations include:
\begin{itemize} 
    \item {
    LCM-LAI (w/o AIA): To show the importance of our article-intervened attention (\textbf{AIA}) mechanism, the first variation model we build is the LCM-LAI without AIA, i.e., using only 
    $\mathbf{X}_f = \mathbf{X}^{(S)} \oplus \mathbf{X}^{(L)}$ and $\mathbf{Y}_f = \mathbf{Y}^{(S)} \oplus \mathbf{Y}^{(L)}$ to predict the matching level labels.
    }
    \item {
    LCM-LAI (w/o LIM): To evaluate the effectiveness of our dependent multi-task learning framework, especially the article prediction sub-task, we build the second variation that LCM-LAI without the legal interaction module (LIM), records as LCM-LAI (w/o LIM).
    This variation only uses the semantic interaction representations, i.e., $\mathbf{X}_f = \mathbf{X}^{(S)}$ and $\mathbf{Y}_f = \mathbf{Y}^{(S)}$ to predict the matching level labels, without optimizing the law article sub-task during training.
    }

    \item {
    LCM-LAI (w/o BIM): To compare whether semantic or legal-rational interactions are more critical in LCM \& LCR tasks, we also design a third variant that LCM-LAI without basic interaction module (BIM), records as LCM-LAI (w/o BIM) or LCM-LAI (only LIM).
    This variation uses only the legal-rational interaction representations, i.e., $\mathbf{X}_f = \mathbf{X}^{(L)} \oplus \mathbf{X}^{(A)}$ and $\mathbf{Y}_f = \mathbf{Y}^{(L)} \oplus \mathbf{Y}^{(A)}$ to predict the matching level labels.
    }
    \item{
    LCM-LAI (LIM without AIA): To further explore which is more effective in LIM between the legal interaction representation (i.e.,$\mathbf{X}^{(L)}$ and $\mathbf{Y}^{(L)}$) and the article-intervened representation (i.e., $\mathbf{X}^{(A)}$ and $\mathbf{Y}^{(A)}$), we construct another variant that LCM-LAI (LIM without AIA), which only uses the legal-rational representation without law intervention to make predictions, i.e., $\mathbf{X}_f = \mathbf{X}^{(L)}$ and $\mathbf{Y}_f = \mathbf{Y}^{(L)}$.
    }
    \item {
    LCM-LAI (only AIA): To further demonstrate that AIA complements the information of legal-rational interaction, we propose the final variation that LCM-LAI (only AIA), which uses the article-intervened legal-rational representation for prediction, i.e., $\mathbf{X}_f = \mathbf{X}^{(A)}$ and $\mathbf{Y}_f = \mathbf{Y}^{(A)}$.
    }
\end{itemize}

\begin{table}[t]
\centering
\caption{Ablation study of LCM-LAI on the LeCaRD dataset.}\label{tab: Ablation Study}
\begin{tabular}{lcccccc}
\toprule
 Methods            & Accuracy      & Macro-Precision       & Macro-Recall      & F1         \\
 \midrule
 LCM-LAI            & $\textbf{73.99}$       & $\textbf{73.60}$       & $\textbf{76.50}$       & $\textbf{74.28}$     \\

 LCM-LAI (w/o AIA)  & $73.07$       & $73.16$       & $75.60$       & $73.70$     \\

 LCM-LAI (w/o LIM)  & $62.54$       & $64.09$       & $63.25$       & $63.16$     \\

 LCM-LAI (w/o BIM)  & $72.45$       & $73.06$       & $74.03$       & $72.43$     \\
 
 LCM-LAI (only AIA) & $71.52$       & $71.74$       & $72.11$       & $71.87$     \\

 LCM-LAI (LIM without AIA) & $ 71.21 $       & $ 72.81 $       & $ 71.13 $       & $ 71.82 $     \\
\bottomrule
\end{tabular}%
\end{table}

Tab.~\ref{tab: Ablation Study} shows the experiment results. It demonstrates that all AIA, LIM, and BIM are necessary for our LCM-LAI to improve performance.
We summarize some other specific observations:
\begin{enumerate}
    \item {
    Compared to other variations, the LCM-LAI (w/o LIM) variation exhibits a considerable decrease in performance, indicating that the law prediction subtask is crucial in enhancing the performance of similar case matching. 
    This further highlights the superior effectiveness of its dependent multi-task learning framework.
    }
    \item {
    Compared to the LCM-LAI (w/o LIM), the improved performances of LCM-LAI (only AIA) and LCM-LAI (LIM without AIA) provide further evidence supporting the higher effectiveness of legal-rational interaction representations over pure semantic interaction representation in the context of similar case matching. 
    This finding again reinforces that the definition of similarity within the task of similar case matching is domain-specific.
    }
    \item {
    In contrast to the LCM-LAI (w/o BIM), LCM-LAI (w/o AIA) shows better performance,  particularly with a notable improvement of $1.57\%$ on the Macro-Recall metric and $1.27\%$ on the F1 score.
    This outcome underscores the capability of semantic interaction representations, which effectively complements the legal-rational interaction.
    }
\end{enumerate}

\subsubsection{\textbf{Ablation of the Legal Correlation Matrix}}
In addition, we try to validate further the efficacy of the legal correlation matrix proposed within the LCM-LAI framework from two distinct perspectives: (1) The successful acquisition of effective alignment relationships among case sentences; (2) The superiority of measuring legal correlation from a legal distribution perspective as opposed to the direct computation of sentence embedding correlations. 
Thus, we retain the integrity of the overall framework, only replace the legal correlation matrix, and generate the following variants:

\begin{itemize}
    \item {
    LCM-LAI (unit) and LCM-LAI (random): To ascertain the efficacy of the proposed legal correlation matrix $\mathbf{C}^{(L)}$ in learning meaningful sentence alignments between cases, we have developed two variant models by substituting $\mathbf{C}^{(L)}$ with a random matrix and a unit matrix, referred to as LCM-LAI (unit) and LCM-LAI (random), respectively.
    }
    \item {
    LCM-LAI (embedding distance): To further substantiate the superiority of assessing legal correlation from the perspective of law distribution, we construct another variant that directly computes cosine similarity between sentences' value vectors as legal-rational relevance between sentences, denoted as LCM-LAI (embedding distance).
    }
\end{itemize}
The results\footnote{All experimental results obtained experimentally on the fifth data division in our 5-fold setting.} are presented in Fig.~\ref{fig: Legal_matrix}. 
In a word, LCM-LAI (embedding distance) performs better than LCM-LAI (unit) and LCM-LAI (unit) performs better than LCM-LAI (random).
LCM-LAI performs best.
By analyzing the results, we can draw the following conclusions:

\begin{figure*}[t]
\centering
  \subfigure[\label{fig: Legal_matrix_acc}]{
    \adjustbox{valign=b}{
      \includegraphics[width=0.5\textwidth]{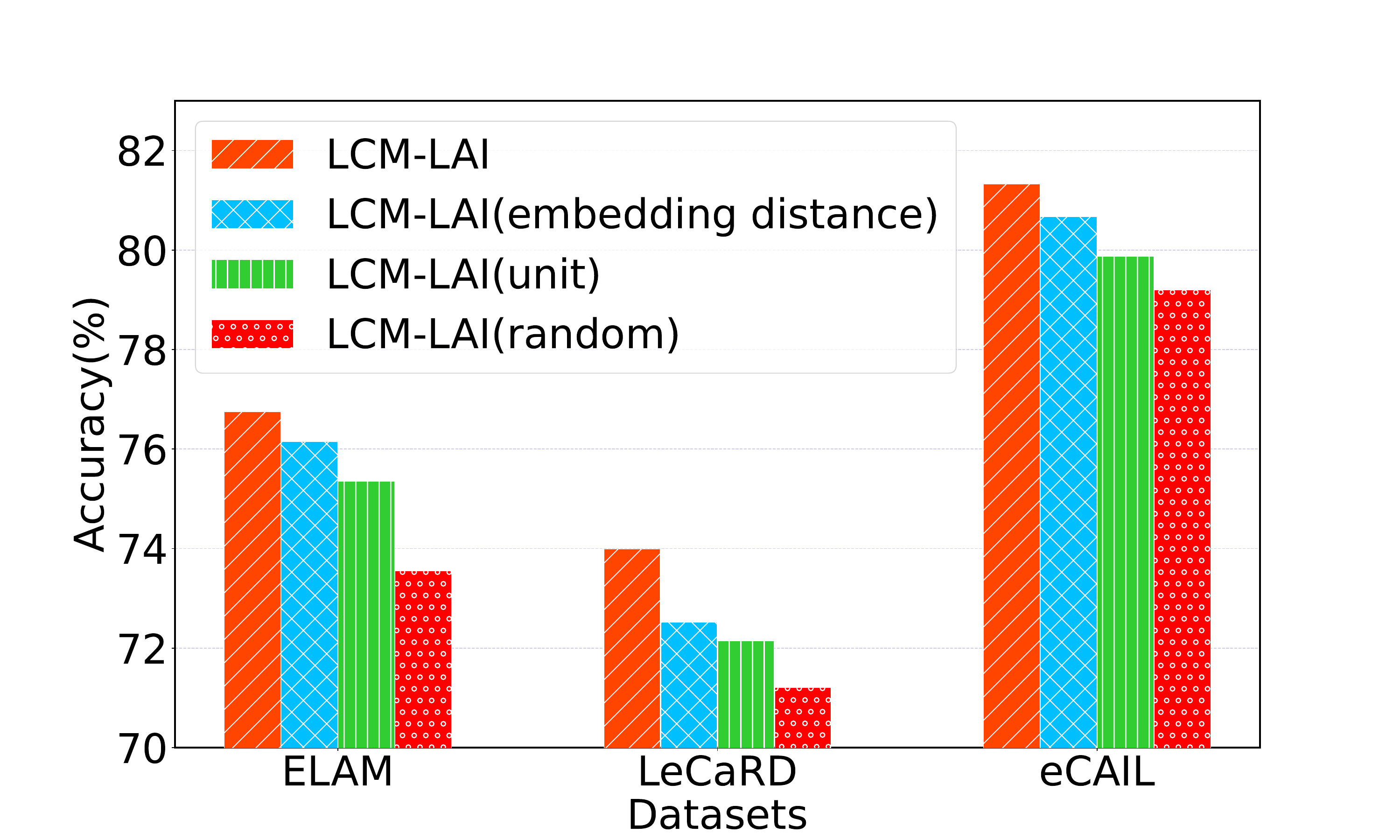}
    }}
  \hspace{-8mm}
  \subfigure[\label{fig: Legal_matrix_f1}]{
    \adjustbox{valign=b}{
      \includegraphics[width=0.5\textwidth]{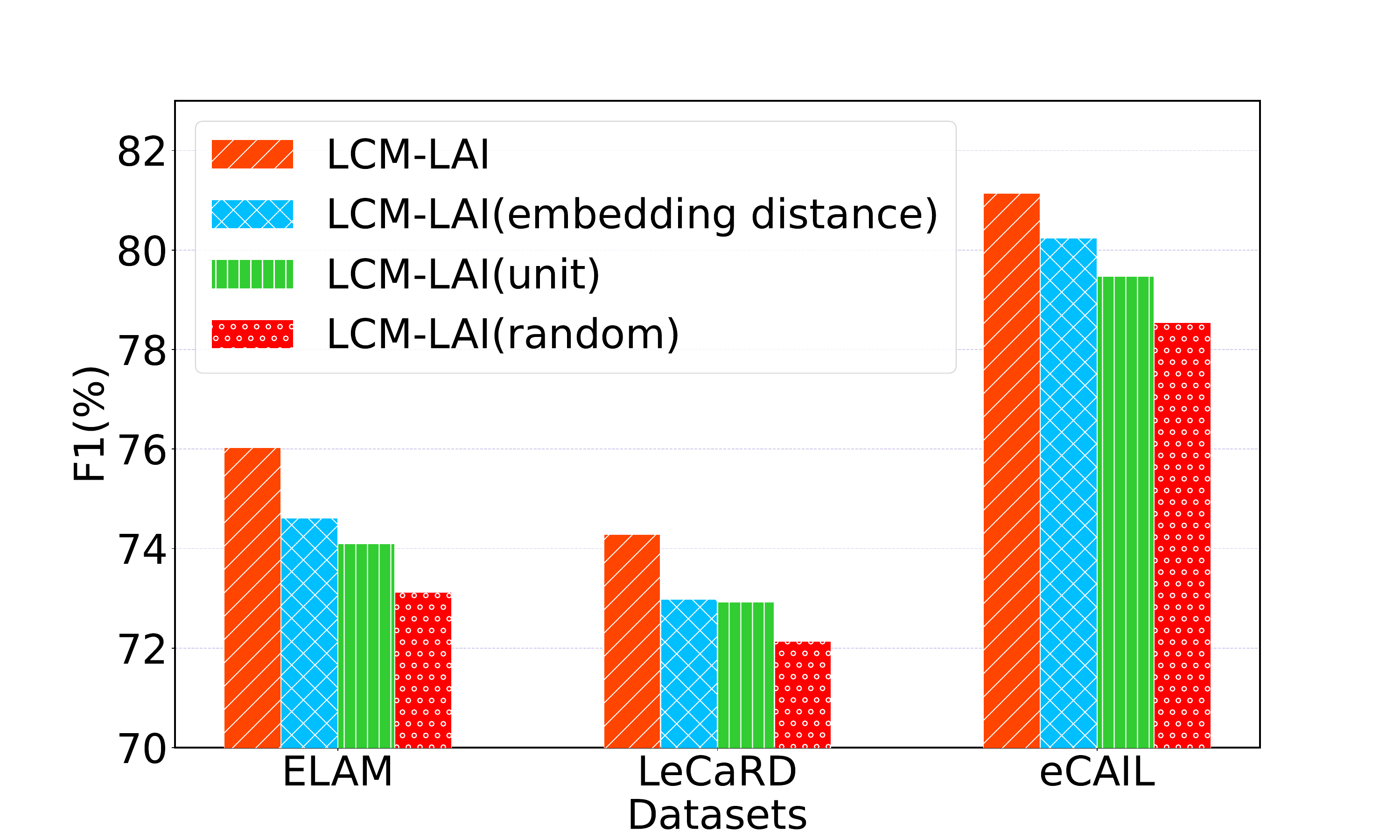}
    }}
\caption{The performance of LCM-LAI w.r.t. different mode of legal correlation matrix on LCM task.
}
\label{fig: Legal_matrix}
\end{figure*}

\begin{enumerate}
    \item{
    The enhanced performance of the original model LCM-LAI and the variant LCM-LAI (embedding distance) in comparison to variants LCM-LAI (unit) and LCM-LAI (random) validate that the legal correlation matrix effectively captures meaningful sentence alignments between cases.
    More fine-grained experiments can be found in Sec \ref{sec:interpretability}.
    }
    \item{
    The superior performance of the original model LCM-LAI relative to the variant LCM-LAI (embedding distance) confirms that assessing the legal-rational correlation between across-case sentences from the perspective of law distribution is more effective than directly computing the similarity between their embedding representations.
    }
    \item{
    The variant LCM-LAI (unit) performs better than the variant LCM-LAI (random).
    We conjecture this is because the standardization of legal document writing causes key circumstances, constitutive elements of crime, and focus of dispute to have specific positional distributions, which makes the sentence alignment relationship matrix between similar cases closer to the unit matrix compared to a random matrix.
    }
\end{enumerate}

\subsection{Inference Efficiency Evaluation (RQ3)}
In the practical retrieval system, inference efficiency is a critical indicator, as it directly impacts the query delay, consequently affecting their overall product experience.
Thus, we examine the inference efficiency by re-ranking the top-k result given by a BOW retrieval model (e.g., BM25), the most typical setting for testing and deploying dense retrieval models.
We conduct this experiment on the LeCaRD dataset.
Regarding our metrics, we report two commonly used efficiency indicators: the re-ranking latency of the top 100 candidates and the FLOPs per query.
Besides, all benchmarks are measured on a single $32$ GB Tesla V$100$ GPU.
For a fair evaluation, we exclude CPU-based text pre-processing and present the primary GPU computation latency since it is a more time-consuming aspect.
To simulate real-world application scenarios, all methods aim to precompute as much of the processing (e.g., the encoding of candidate cases) as possible offline and parallelize the remaining online computations. 
As a result, only a negligible cost is incurred.
As for the FLOPs per query, we estimate each model with the torchprofile\footnote{\url{htps://github.com/mit-han-lab/torchprofle}} library.

Fig.~\ref{fig: Time Efficiency} show the experiment results.
As we can see, LCM-LAI maintains a reasonable range of re-ranking latency and FLOPs per query.
Specifically, we get some summing-up as follows:

\begin{figure*}[t]
\centering
  \subfigure[\label{fig: Latency}]{
    \adjustbox{valign=b}{
      \includegraphics[width=0.465\textwidth]{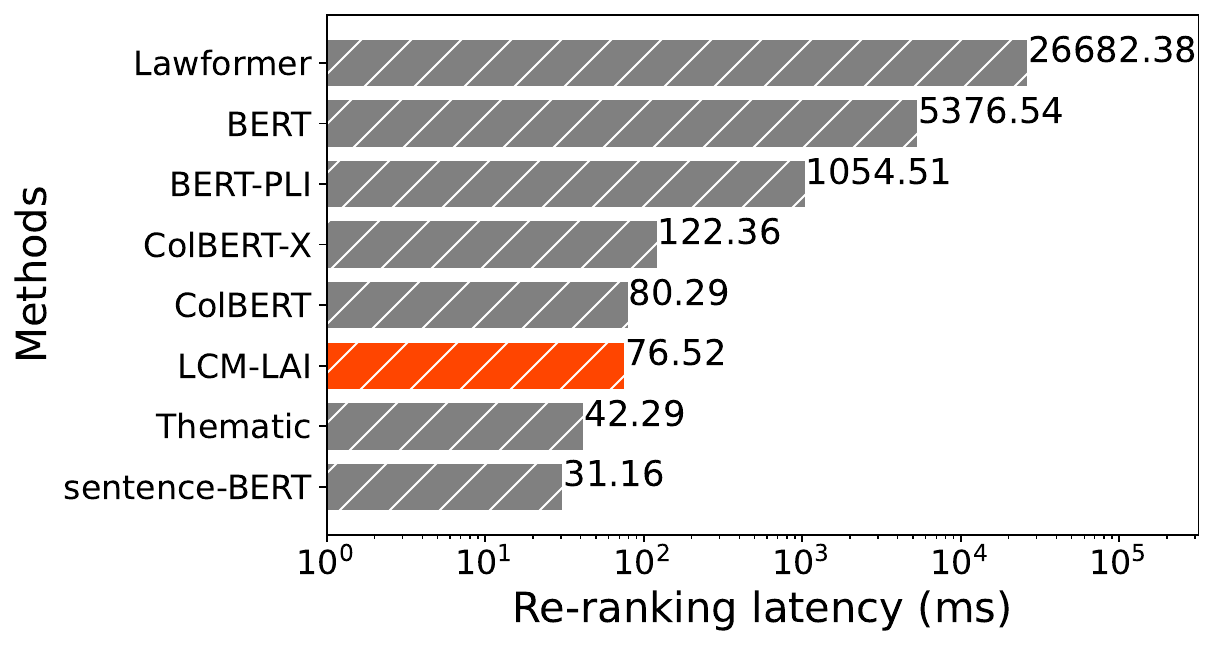}
    }}
  \hspace{5mm}
  \subfigure[\label{fig: FLOPs}]{
    \adjustbox{valign=b}{
      \includegraphics[width=0.43\textwidth]{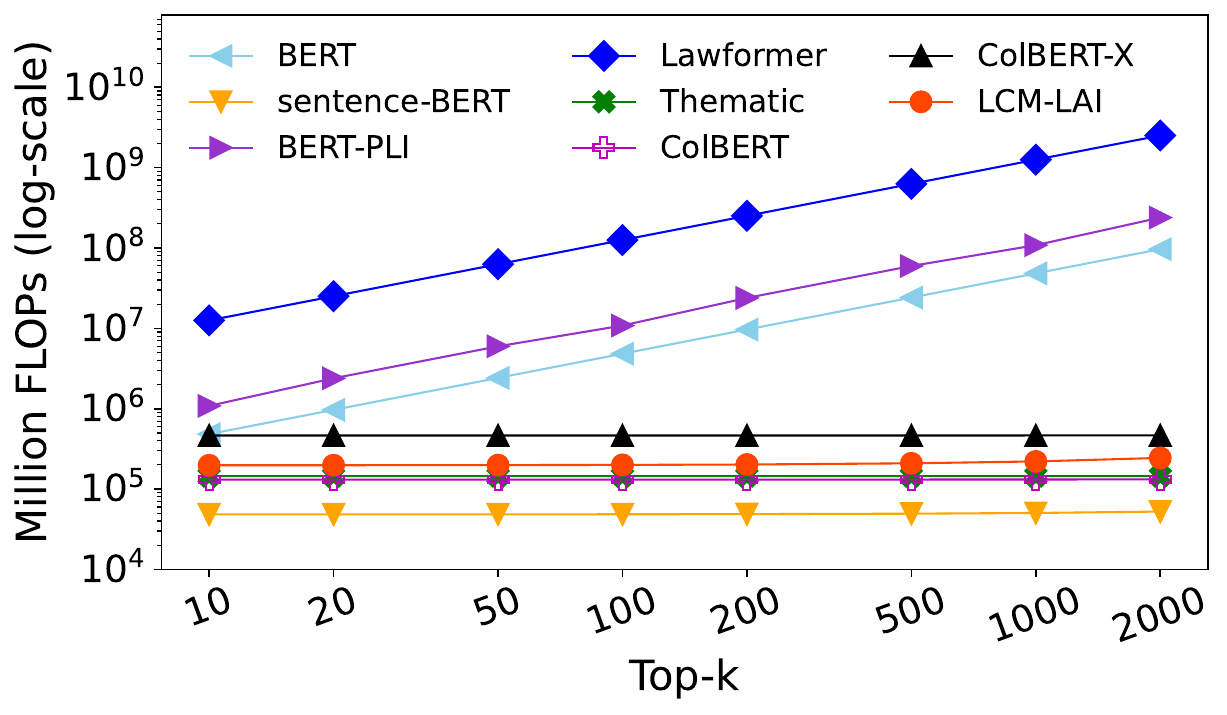}
    }}
\caption{The performance of efficiency evaluation.
(a) Time latency under the top-100 re-ranking setting.
(b) FLOPs (in millions) per query of the re-ranking depth $k$.
To make an even distribution, we use logarithmic coordinates for sub-figure (b) abscissa.
}
\label{fig: Time Efficiency}
\end{figure*}

\begin{enumerate}
    \item {
    Referring to Fig.~\ref{fig: Latency}, under the top-100 re-ranking setting, LCM-LAI's inference latency of $76.52ms$ effectively meets the response time criteria set by the online retrieval system ($<<500ms$). Meanwhile, it's considerably lower than other all-to-all interaction baselines (i.e., BERT, BERT-PLI, and Lawformer) that model the interactions between words within as well as across query and candidates at the same time.
    }
    \item {
    Analyzing Fig.~\ref{fig: FLOPs}, we note that the FLOPs of all-to-all interaction baselines (i.e., Lawformer, BERT, and BERT-PLI) grow linearly with the number of candidate sets.
    This is because their interaction patterns must be recalculated for every case pair, which is impossible to precompute offline.
    On the contrary, the FLOPs of LCM-LAI do not change as the number of candidate sets grows.
    This is attributed to the late interaction framework of LCM-LAI, which allows us offline to precompute the computationally expensive Sentence Embedding module and only online compute the cheap interaction modules (i.e., LIM and BIM).
    Specifically, taking the legal case $Y$ as a candidate, the majority of computations, including $\mathbf{y}_i$, $\mathbf{L}_k^Y$, $\Lambda^{Y}$, and $\mathbf{v}_j^Y$, can be conducted offline.
    Hence, when the candidate set is expanded, the computational overhead of LCM-LAI can remain consistent, similar to representation-focused rankers such as Sentence-Bert and Thematic.
    }
\end{enumerate}

\subsection{The Interpretability of Matching (RQ4)}\label{sec:interpretability}
Interpretability assumes a crucial role by fulfilling two fundamental objectives in the legal domain. 
Firstly, it effectively disseminates accurate knowledge about the law to the public through reasonable explanations. 
Secondly, it assists judges in their decision-making process by offering professional argument logic.
Since our LCM-LAI uses considerable interactive operations and attention mechanisms, we expect the model to have some interpretability.
We separately visualize the semantic correlation matrix of BIM, the legal correlation matrix of LIM, and the attention matrix of AIA for an example legal case pair from the ELAM test set.
To evaluate the overall quality, we also compare our visualization with the ground truth obtained from the alignment labels  (cf. Fig.~\ref{fig: HOT_truth}).

Fig.~\ref{fig: Interpretability} shows the results, where the darker block is, the higher the relation score the corresponding sentence pair gets. 
Overall, for all three gold alignment labels, i.e., ($S_{11}^{1}$, $S_{6}^{2}$), ($S_{8}^{1}$, $S_{8}^{2}$), and ($S_{11}^{1}$, $S_{11}^{2}$), all interaction matrices and attention matrices can sparsely cover two of them, i.e., ($S_{8}^{1}$, $S_{8}^{2}$) and ($S_{11}^{1}$, $S_{11}^{2}$), which proves that LCM-LAI has some interpretability.
Then, referring to specific sentence descriptions with high scores (cf. Fig.~\ref{fig: explain_sentences}), we find the following interesting observations:

\begin{figure*}[t]
\centering
 \subfigure[\label{fig: HOT_truth}]{\includegraphics[width=0.245\linewidth]{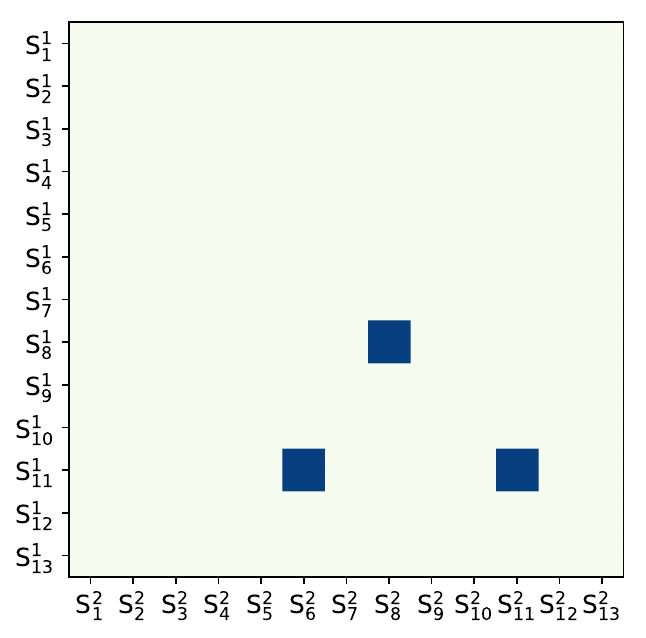}}
 \subfigure[\label{fig: HOT_BIM}]{\includegraphics[width=0.245\linewidth]{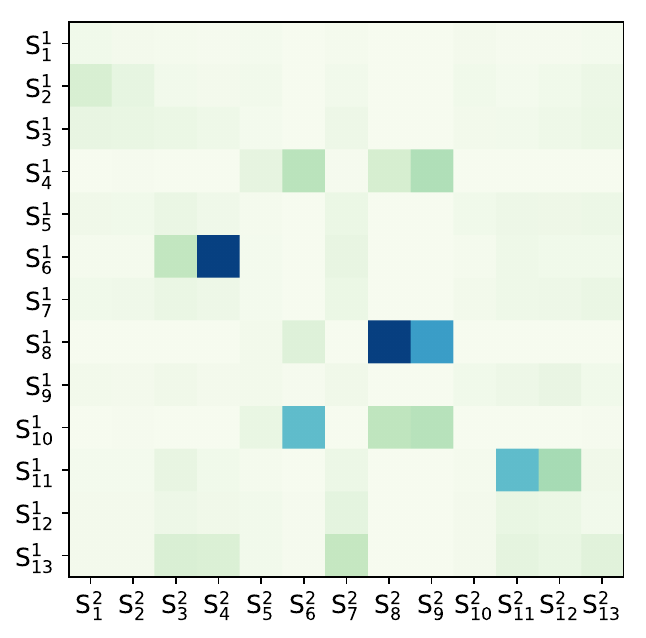}}
 \subfigure[\label{fig: HOT_LIM}]{\includegraphics[width=0.245\linewidth]{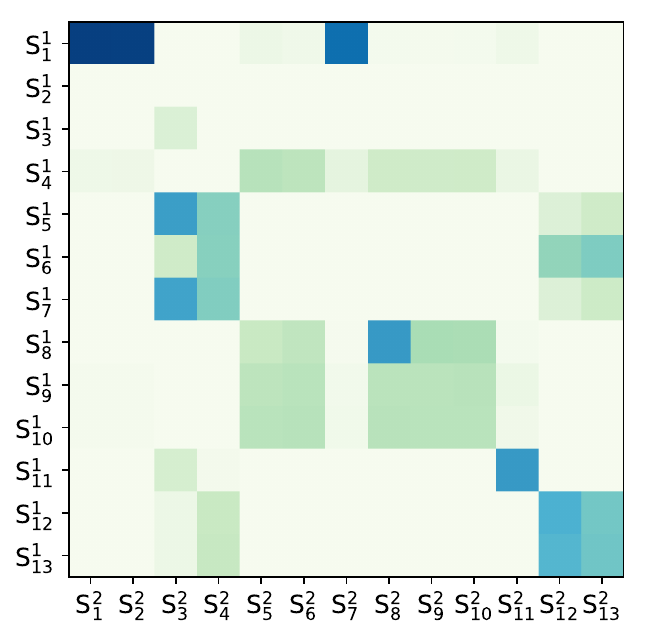}}
 \subfigure[\label{fig: HOT_AIA}]{\includegraphics[width=0.245\linewidth]{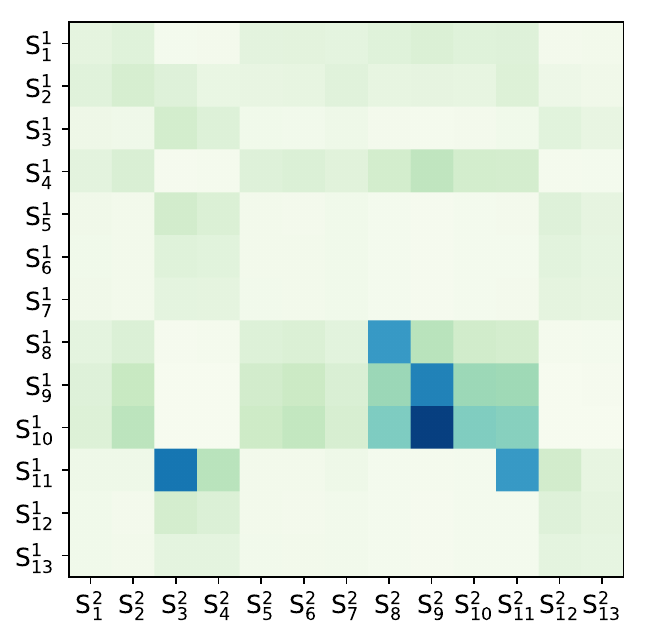}}
 \subfigure[\label{fig: explain_sentences}]{\includegraphics[width=\linewidth]{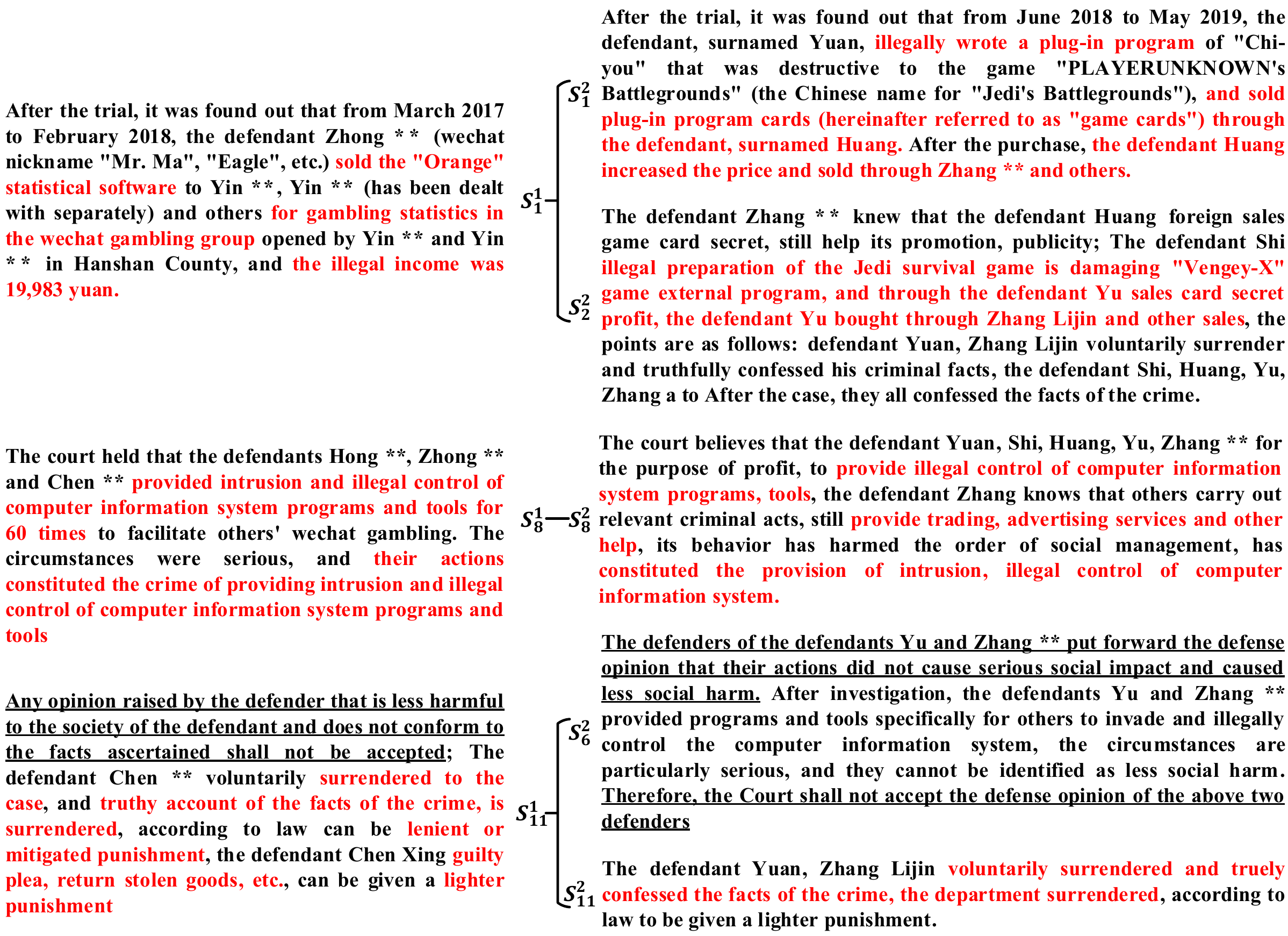}}
\caption{The visualization of sentence alignment for an example in the ELAM test set.
(a) The human-labeled alignments;
(b), (c), and (d) The attention score visualization for semantic correlation matrix (cf. Sec~\ref{subsubsec: semantic correlation matrix}), legal correlation matrix (cf. Sec~\ref{subsubsec: law correlation matrix}), and legal correlation matrix combined with AIA (cf. Sec~\ref{subsubsec: law interaction encoder}); 
(e) The sentence pairs with ground-truth alignment labels or the highest interaction scores.
}
\label{fig: Interpretability}
\end{figure*}

\begin{enumerate}
    \item {
    We notice that the legal correlation matrix captures some sentence pairs far away from the gold alignment labels, i.e., ($S_{1}^{1}$, $S_{1}^{2}$) or ($S_{1}^{1}$, $S_{2}^{2}$) when other semantic correlation matrix and AIA centrally capture the matching relations around the aligned labels.
    From Fig.~\ref{fig: explain_sentences}, we find that these sentence pairs capture the interaction information about the key circumstance of fact description instead of the constitutive elements of the crime (represented by($S_{8}^{1}$, $S_{8}^{2}$) and ($S_{11}^{1}$, $S_{11}^{2}$)), or the focus of disputes (represented by ($S_{11}^{1}$, $S_{6}^{2}$)).
    This notable observation offers a conceivable rationale for the remarkable efficacy demonstrated by the LCM-LAI model on the LeCaRD dataset. 
    In the pre-processing phase, a conscious decision is made to retain solely the factual components of the case, eliminating court summary text that includes inferences and points of dispute. Consequently, the models can predict matching labels solely by capturing the interactive information in the case facts, which aligns perfectly with the strengths of the LCM-LAI model.
    }
    
    \item {
    We also notice that all three interaction matrices can hardly capture the labeled sentence pair ($S_{11}^{1}$, $S_{6}^{2}$), which reflects that LCM-LAI is still difficult to detect the controversial focus of a legal case (i.e., the underlined part of Fig.~\ref{fig: explain_sentences}).
    The focus of the dispute usually consists of a defense statement and a court decision, with the former often pleading for reduced charges or penalties.
    However, these statements are commonly deemed as noise or irrelevant to the law prediction subtask, which makes it difficult for the LCM-LAI model to pay attention to the focus of the dispute. 
    This is also the shortcoming of the LCM-LAI and inspires future enhancements to it.
    }
\end{enumerate}

\begin{table*}[t]
\centering
\small
\caption{
The matching performance on the LeCaRD dataset with different backbones. The results of our LCM-LAI are in \textbf{bold}. $\dagger$ denotes LCM-LAI achieves significant improvements over all existing baselines in paired t-test with $p$-value $\textless 0.05$.
}\label{tab: Different-backbones}
\begin{tabular}{lcccc}
\toprule
Methods   & Accuracy   & Macro-Precision   & Macro-Recall    & F1\\
 \midrule

BERT          
                    & $ 65.02 $     & $ 65.23 $     & $ 63.50 $     & $ 64.26 $     \\

$\textbf{LCM-LAI}_{BERT}$    
                    & $ \textbf{73.99}^{\dagger} $ & $ \textbf{73.60}^{\dagger} $ 
                    & $ \textbf{76.50}^{\dagger} $ & $ \textbf{74.28}^{\dagger} $\\

\midrule
Lawformer          
                    & $ 65.63 $     & $ 64.81 $     & $ 65.73 $     & $ 65.17 $\\
$\textbf{LCM-LAI}_{Lawformer}$    
                    & $ \textbf{73.37}^{\dagger} $ & $ \textbf{75.51}^{\dagger} $ 
                    & $ \textbf{73.25}^{\dagger} $ & $ \textbf{73.82}^{\dagger} $\\
\midrule
Legal-RoBERTa         
                    & $ 66.87 $     & $ 65.01 $  & $ 65.60 $     & $ 65.28 $\\
$\textbf{LCM-LAI}_{Legal-RoBERTa}$   
                    & $ \textbf{74.30}^{\dagger} $ & $ \textbf{76.67}^{\dagger} $ 
                    & $ \textbf{73.90}^{\dagger} $ & $ \textbf{74.66}^{\dagger} $\\
\midrule
XLNet         
                    & $ 64.09 $     & $ 63.93 $   & $ 63.16 $     & $ 63.41 $\\
$\textbf{LCM-LAI}_{XLNet}$   
                    & $ \textbf{71.83}^{\dagger} $ & $ \textbf{72.46}^{\dagger} $ 
                    & $ \textbf{72.33}^{\dagger} $ & $ \textbf{72.32}^{\dagger} $\\

\midrule
BGE         
                    & $ 64.71 $     & $ 64.15 $   & $ 64.84 $     & $ 64.43 $\\
$\textbf{LCM-LAI}_{BGE}$   
                    & $ \textbf{72.76}^{\dagger} $ & $ \textbf{73.97}^{\dagger} $ 
                    & $ \textbf{72.61}^{\dagger} $ & $ \textbf{73.19}^{\dagger} $\\
\bottomrule
\end{tabular}%
\end{table*}

\subsection{Robustness about Different Backbones}

As the proposed framework of LCM-LAI is backbone-independent, it is easy to transfer across different backbone models.
In this section, to demonstrate that the LCM-LAI framework can improve different backbone models, we conducted a comparative experiment on the fifth iteration of the LeCaRD dataset.
In addition to the BERT and Lawformer used in the previous experiments, we also compare the following two popular PLMs:

\begin{itemize}
\item \textbf{Legal-RoBERTa}~\cite{xiao2021Lawformer}\footnote{\url{https://huggingface.co/IDEA-CCNL/Erlangshen-Roberta-330M-Similarity}}: the RoBERTa~\cite{liu2019roberta} model fine-tuned by Chinese legal corpus based on the RoBERTa-wwm-ext~\cite{cui2021pre} checkpoint. 
This model follows the whole word masking strategy, in which the tokens that
belong to the same word will be masked simultaneously.

\item \textbf{XLNet}~\cite{NEURIPS@XL_Net}\footnote{\url{https://huggingface.co/hfl/chinese-xlnet-base}}:
a PLM model that employs a generalized autoregressive pertaining strategy. 
It utilizes a different architecture and extra tokens compared to BERT.
By integrating the segment recurrence mechanism and relative encoding scheme of Transformer-XL into its perturbation process, XLNet demonstrates superior performance on tasks involving longer text sequences compared to BERT.

\item \textbf{BGE}~\cite{xiao2023BGE}\footnote{\url{https://huggingface.co/BAAI/bge-large-zh-v1.5}}: 
a general embedding model that is pre-trained using RetroMAE~\cite{xiao2022retromae} and trained on large-scale pair data using contrastive learning.
Compared with the general PLM, this model has proved to be more suitable for document retrieval tasks.

\end{itemize}
Note that for all original backbone models compared in this experiment, we all take the case document pair as input and utilize the fully interactive architecture, as it performs better than the Sentence-BERT-like architecture.
When the different bases are coupled with our LCM-LAI architecture, the experimental setup used is also completely consistent with that described in Sec.~\ref{sec: hyperparametter_setting}.

The experiment results are presented in Tab. \ref{tab: Different-backbones}. 
From the observations, we can draw the following conclusions:

\begin{enumerate}
    \item{
    For arbitrary backbone models, our LCM-LAI framework achieves significant improvement ranging from $7.43\%$ to $13.00\%$, further demonstrating the robustness of the proposed model.
    }
    \item{
    We observe that BERT, Lawformer, and Legal-RoBERTa perform better than XLNet and BGE for the original backbone models.
    We contend that this is because the former models have been pre-trained with a professional corpus in the judicial domain,  making them more adept at comprehending legal terminology.
    }
\end{enumerate}


\section{Discussion 1: Whether LCM-LAI benefits from fine-grained legal-rational and alignment labels?}

As mentioned in Sec.~\ref{sec: basic_performance}, we attribute the performance advantage of IOT-match over our LCM-LAI on the eCAIL dataset to the additional annotations they used.
In this section, we conduct additional experiments to discuss whether our LCM-LAI model is similarly able to benefit from these additional annotations.
IOT-match uses three kinds of additional annotations, including rationale labels for each sentence, the binary alignment matrix for pairwise cases, and explanatory sentences that describe matching relationships between cases.
Since the last additional annotation is employed to train a generative model to generate interpretable statements and cannot be naturally combined with our LCM-LAI,
only the former two are chosen in this experiment to enhance our LCM-LAI model.
Here we give the definitions of these two additional annotations according to the original IOT-Match~\cite{yu2022Explainable}.
At first, we denote the rationale labels of case $X$ and $Y$ as $\mathbf{r}^{X}$ and $\mathbf{r}^{Y}$.
The rationale labels are associated with the sentence embeddings, 
i.e., $\mathbf{r}^{X}=\{r_{x_i}\}_{i=1}^{n_x}$ and $\mathbf{r}^{Y}=\{r_{y_j}\}_{j=1}^{n_y}$, where the rationale label of a sentence $x_i$ is designed following:
$$
r_{x_{i}} =
\begin{cases}
0 & x_{i} \text{ is not a rationale}, \\
1 & x_{i} \text{ is a key circumstance}, \\
2 & x_{i} \text{ is a constitutive element of crime}, \\
3 & x_{i} \text{ is a focus of disputes}. \\
\end{cases}
$$
Then, we denote the binary alignment matrix between case $X$ and $Y$ as $\mathbf{A}=[a_{i,j}]\in \{0, 1\}^{n_x \times n_y}$, in which each element $a_{i,j}$ is annotated manually by the following formula,
$$
a_{i, j} =
\begin{cases}
0 & r_{x_i} \neq r_{y_j}, \\
1 & r_{x_i} = r_{y_j} \ \& \ x_i \cong y_j. \\
\end{cases}
$$
where $x_i \cong y_j$ means the sentences corresponding to $x_i$ and $y_j$
are semantically-similar. $a_{i,j}= 1$ means aligned rationales while
$a_{i,j}=0$ means misaligned rationales.

Utilizing additional annotations, we focus on improving the LIM of LCM-LAI, as this module primarily extracts legal-rational information and the legal-rational interaction information between cases, which aligns precisely with the focus of the additional annotations.
On the one hand, for the rationale label, taking case $X$ as an example, we use the value vector $\mathbf{v}_{i}^{X}$ (c.f., Eq.\ref{eq: sub_attention_sum} and Eq.\ref{eq: legal_inter_rep}) of each sentence $x_i$ to predict the corresponding rationale label.
Because LCM-LAI directly weighted sums value vectors of all sentences to predict applicable law labels, i.e., 
value vectors directly reflect the legal-rational information extracted by LCM-LAI.
As for the specific implementation, we use a simple linear classifier to predict the rationale labels, whose formula is shown as follows:
\[
\{ P(r=t | x_i) \}_{t=0}^{3} = \text{softmax}(\mathbf{W}_{r}\mathbf{v}_i^X + \mathbf{b}_{r} )
\]
where the softmax converts a 4-dimensional vector to a distribution over four rationale classes, $\mathbf{W}_{r}$ and $\mathbf{b}_{r}$ are trainable parameters.
We select the rationale class with the max probability as our prediction, i.e., 
$\hat{r}_{x_i} = \arg \max_{t\in \{0, \cdots, 3\}}P\left ( r=t| x_i\right)$.
Thus, following the practice of IOT-Match~\cite{yu2022Explainable}, for the case pair containing $X$ and $Y$, we also select the cross-entropy loss between the ground-truth rationale labels of each sentence and the corresponding predictions as the rationale identification loss $\mathscr{L}_{rationale}$.
The specific formula is as follows,
\[
    \mathscr{L}_{rationale} = \sum_{t=0}^{3}\left( 
    \sum_{i=1}^{n_x} \delta(r_{x_i}, t) \log \left(P (\hat{r}_{x_i}=t | x_{i}) \right)
    +  \sum_{j=1}^{n_y} \delta(r_{y_j}, t) \log \left(P(\hat{r}_{y_j}=t | y_{j})\right)
    \right),
\]
where $\delta(r, t) = 1$ if $r = t$ else 0. 
On the other hand, as the legal correlation matrix $\mathbf{C}^{(L)}$ is designed to capture the legal interaction information between cases, we use the label of the alignment matrix $\mathbf{A}$ to constrain it.
Following the IOT-Match, we also treat the KL divergence between $\mathbf{A}$ and $\mathbf{C}^{(L)}$ as the objective function.
Its formula is as follows,
\[
    \mathscr{L}_{align} = \text{KL}( \mathbf{A} || \mathbf{C}^{(L)}) = \sum_{i, j} a_{i,j} \log \frac{c_{i, j}^{(L)}}{a_{i,j}}.
\]
Thus, by adding different enhancing objective functions, we get the following variants:
\begin{itemize}
\item \textbf{LCM-LAI(+ rationale)}: the variant enhanced by using rationale labels, of which the overall loss function is $\mathscr{L} = \mathscr{L}_{a} + \mathscr{L}_{m} + \mathscr{L}_{rationale}$.

\item \textbf{LCM-LAI(+ align)}: the variant enhanced by using alignment matrix labels, of which the overall loss function is  $\mathscr{L} = \mathscr{L}_{a} + \mathscr{L}_{m} + \mathscr{L}_{align}$.

\item \textbf{LCM-LAI(+ both)}: the variant enhanced by using both alignment matrix and rationale labels, of which the overall loss function is  $\mathscr{L} = \mathscr{L}_{a} + \mathscr{L}_{m} + \mathscr{L}_{rationale} + \mathscr{L}_{align}$.
\end{itemize}

According to the experimental results shown in Tab.~\ref{tab: additional annotation},
we can get the following conclusions:

\begin{enumerate}
    \item{
    With the help of the additional labels, LCM-LAI(+ both) achieves better performance than IOT-Match, with Accuracy, Macro-Precision, Macro-Recall, and F1 exceeding $1.24\%$, $1.07\%$, $1.66\%$ and $1.54\%$ respectively.
    }
    \item{
    Both the rationale labels and the align matrix labels indeed improve the performance of our LCM-LAI.
    }
\end{enumerate}

\begin{table*}[t]
\centering
\small
\caption{
The matching performance of LCM-LAI on the eCAIL dataset with additional annotations.
}\label{tab: additional annotation}
\begin{tabular}{lcccc}
\toprule
Methods   & Accuracy   & Macro-Precision   & Macro-Recall    & F1\\
 \midrule

IOT-Match          
                    & $ 82.00 $     & $ 82.10 $     & $ 81.92 $     & $ 81.90 $     \\

$\textbf{LCM-LAI}$    
                    & $ 81.07 $ & $ 81.01 $ 
                    & $ 81.23 $ & $ 80.89 $\\
 
$\textbf{LCM-LAI(+ rationale)}$    
                    & $ 82.05 $ & $ 82.03 $ 
                    & $ 82.41 $ & $ 82.29 $\\

$\textbf{LCM-LAI(+ align)}$    
                    & $ 82.40 $ & $ 81.91 $ 
                    & $ 82.18 $ & $ 81.95 $\\

$\textbf{LCM-LAI(+ both)}$    
                    & $ \textbf{83.24} $ & $ \textbf{83.07} $ 
                    & $ \textbf{83.58} $ & $ \textbf{83.44} $\\
\bottomrule
\end{tabular}%
\end{table*}

\section{Discussion 2: Generalizability and Limitations}
\noindent {\bf Generalizability of LCM-LAI in civil law systems.}
It should be emphasized that compared with the general document retrieval model, the proposed LCM-LAI only differs in the requirement to take the text definition of law articles as inputs and the law article prediction sub-task to capture the legal-rational correlation between cases. The applicability of LCM-LAI is extensive within civil law systems (e.g., China, France, Japan, and so on), primarily due to the explicit codification of law articles by the relevant judicial authorities. There are two reasons why this paper only conducts experiments on Chinese legal data: 1) The Chinese legal system can effectively represent the civil law system; 2) For authors, the accessibility of Chinese legal data is the least difficult, especially the detailed text definition of relevant law articles.

\noindent {\bf Limitations of LCM-LAI in common law systems.}
Due to the lack of written law articles, directly applying the proposed LCM-LAI model in the common law system (e.g., Britain, America, Canada, etc.) is limited.
Here, we also provide an envisioned feasible strategy to apply LCM-LAI to the common law system.
As mentioned in Sec.~\ref{sec:analysis}, LCM-LAI treats the law articles as the summary of the legal-rational information for the corresponding class of cases, an intuitive way to make LCM-LAI work is to summarize the law substitutes from a large number of legal cases.
Fortunately, in both common law systems and civil law systems, it is natural to cluster historical legal cases according to the labels of the convictions.
Next, by applying text summarization, topic modeling, and even ChatGPT technology, we can easily derive a text summarization for each set of legal cases, which can substitute for the law articles in our LCM-LAI.
Of course, the quality of this summary generation will affect the effect of the subsequent use of LCM-LAI.
As for the specific effects of the above strategy, we will further explore it in the future.



\section{Conclusion and Future Work} \label{sec:conclusion}
In this paper, we propose an end-to-end model, LCM-LAI, to solve both the LCR and LCM tasks.
On the one hand, LCM-LAI designs a dependent multi-task learning framework to solve the challenge of extracting legal-rational information of cases by adding a sub-task of law article prediction.
Besides, we carry out some theoretical derivation to prove that LCM-LAI is more reasonable than the general multi-task learning framework, which is also confirmed by experiment results.
On the other hand, different from modeling general semantic interaction, LCM-LAI more effectively models the legal-rational correlation between across-case sentences from the perspective of law distribution.
Specifically, LCM-LAI optimizes specifically for the article prediction sub-task with its novel article-aware attention mechanism. 
This feature generates a law attention score vector for each sentence that reflects the law distribution, enabling the computation of legal-rational correlations across case sentences without relying on sentence representations.
Adequate results on two practical tasks on four real-world datasets demonstrate the effectiveness of LCM-LAI.

\begin{acks}
This work was supported in part by the National Key R\&D Program of China (2023YFC3306100), National Natural Science Foundation of China (62372362).

\end{acks}
\bibliographystyle{ACM-Reference-Format}
\bibliography{sample-base}










\end{document}